\documentclass[journal]{IEEEtran}
\usepackage{moreverb,url}
\usepackage[colorlinks,bookmarksopen,bookmarksnumbered,citecolor=red,urlcolor=red]{hyperref}
\usepackage{amsthm}
\usepackage{url}
\usepackage{multicol}
\usepackage{soul}
\usepackage{color}
\usepackage{flushend}
\usepackage{times}
\usepackage{graphicx}
\usepackage{lipsum}
\usepackage{setspace}
\usepackage{epsfig}
\usepackage{mathptmx}
\usepackage{amsmath}
\usepackage{bm}
\usepackage{amssymb}
\usepackage{tabularx}
\usepackage{mathtools}
\usepackage{color,soul}
\usepackage{paralist}
\usepackage{tikz}
\usepackage{float}
\usepackage{pdfpages}
\usepackage{makecell}
\usepackage{array}
\usepackage{xcolor}
\usepackage{caption}
\usepackage{subfigure}
\usepackage{enumerate}
\usepackage{url}
\usepackage{pgfgantt}
\usepackage{amsfonts}
\usepackage{textcomp}
\usepackage{multirow,diagbox}
\usepackage{threeparttable}
\usepackage{booktabs}
\usepackage[noadjust]{cite}

\newcolumntype{L}[1]{>{\raggedright\let\newline\\\arraybackslash\hspace{0pt}}m{#1}}
\newcolumntype{C}[1]{>{\centering\let\newline\\\arraybackslash\hspace{0pt}}m{#1}}
\newcolumntype{R}[1]{>{\raggedleft\let\newline\\\arraybackslash\hspace{0pt}}m{#1}}
\newcommand{\inv}{^{-1}}
\newcommand{\tr}{^{\!\top}}

\newcommand{\argmin}{\operatornamewithlimits{argmin}}

\newcommand{\se}{\mathrm{se}(3)}
\newcommand{\SE}{\mathrm{SE}(3)}

\newcommand{\R} {{\rm I\!R}}
\newcommand{\E} {{\rm I\!E}}

\usepackage{xcolor}  % Needed for colours - load before mdframed

\begin{document}
\title{VDO-SLAM: A Visual Dynamic Object-aware SLAM System}
\author{Jun~Zhang$^{\bf{[co]}}$, Mina~Henein$^{\bf{[co]}}$, Robert~Mahony and~Viorela~Ila% <-this % stops a space
\thanks{Jun Zhang, Mina Henein and Robert Mahony are with the Australian National University (ANU), 0020 Canberra, Australia.
        {\tt \{jun.zhang2,mina.henein,robert.mahony\}@anu.edu.au}}%
\thanks{Viorela Ila is with the University of Sydney (USyd), 2006 Sydney, Australia.
        {\tt viorela.ila@sydney.edu.au}}%
\thanks{[\textbf{co}]: The two authors contributed equally to this work.}%
\thanks{$^{\bf{*}}$\url{https://github.com/halajun/vdo_slam}}}

\markboth{Manuscript Only}% submitted to IEEE Transactions on Robotics
{Shell \MakeLowercase{\textit{et al.}}: Bare Demo of IEEEtran.cls for IEEE Journals}

\maketitle

\begin{abstract}
% Dynamic scene understanding and modelling plays an important role in the deployment of autonomous mobile robotic systems for a wide range of important real world applications, such as planning, control and autonomous driving, etc.
% This paper proposes to decompose the dynamic scene into multiple 6-DoF motions generated via rigid or near rigid bodies (objects), which can be easily integrated into a SLAM framework and solved as a graph optimization problem. 

% The scene rigidity assumption, also known as the static world assumption, is common in SLAM algorithms.
% Such strong assumption limits the deployment of autonomous mobile robotic systems in a wide range of important real world applications involving highly dynamic and unstructured environments.
% Most existing algorithms operating in complex dynamic environments simplify the problem by removing moving objects from consideration or tracking them separately.
% Arguably, the integration of dynamic information into a SLAM framework has potential benefits, including better SLAM performance, as well as wider applications for dynamic scene understanding such as planning, control and autonomous driving, etc.
Combining Simultaneous Localisation and Mapping (SLAM) estimation and dynamic scene modelling can highly benefit robot autonomy in dynamic environments. Robot path planning and obstacle avoidance tasks rely on accurate estimations of the motion of dynamic objects in the scene.
%poses new challenges for SLAM in dynamic environments. 
%Solving this problem becomes practically important for autonomous mobile robotic systems in a wide range of real world applications such as planning, control and obstacle avoidance, etc.
This paper presents VDO-SLAM, a robust visual dynamic object-aware SLAM system that exploits semantic information to enable accurate motion estimation and tracking of dynamic rigid objects in the scene without any prior knowledge of the objects' shape or geometric models.
The proposed approach identifies and tracks the dynamic objects and the static structure in the environment and integrates this information into a unified SLAM framework. This results in highly accurate estimates of the robot's trajectory and the full $\SE$ motion of the objects as well as a spatiotemporal map of the environment.
 %resulting in accurate estimation of the robot and  objects' motion as well as a spatiotemporal map of the environment.
The system is able to extract linear velocity estimates from objects' $\SE$ motion providing an important functionality for navigation in complex dynamic environments.
We demonstrate the performance of the proposed system on a number of real indoor and outdoor datasets and 
the results show consistent and substantial improvements over the state-of-the-art algorithms.
An open-source version of the source code is available$^{\bf{*}}$.

% The scene rigidity assumption, also known as the static world assumption, is common in SLAM algorithms.
% Most existing algorithms operating in complex dynamic environments simplify the problem by removing moving objects from consideration or tracking them separately.
% Such strong assumptions limit the deployment of autonomous mobile robotic systems in a wide range of important real world applications involving highly dynamic and unstructured environments.
% This paper presents VDO-SLAM, a robust object-aware dynamic SLAM system that exploits semantic information to enable motion estimation of rigid objects in the scene without any prior knowledge of the objects shape or motion models.
% The proposed approach integrates dynamic and static structures in the environment into a unified estimation framework resulting in accurate robot pose and spatiotemporal map estimation.
% We provide a way to extract velocity estimates from object pose change of moving objects in the scene providing an important functionality for navigation in complex dynamic environments.
% We demonstrate the performance of the proposed system on a number of real indoor and outdoor datasets.
% Results show consistent and substantial improvements over state-of-the-art algorithms.
% An open-source version of the source code is available$^{\bf{*}}$.
\end{abstract}


% \fi
\IEEEpeerreviewmaketitle

%%%%%%%%%%%%%%%%%%%%%%%%%%%%%%%%%%%%%%%%%%%%%%%%%%%%%%%%%%%%%%%%%%%%%%%%%%%%%%%%%%%%%%%%%%%%%%%%%%%%%%%%%%%%%%%

\section{Introduction}
\label{sec:intro}
\begin{figure}[ht]
 \centering
 \includegraphics[width=1.\columnwidth]{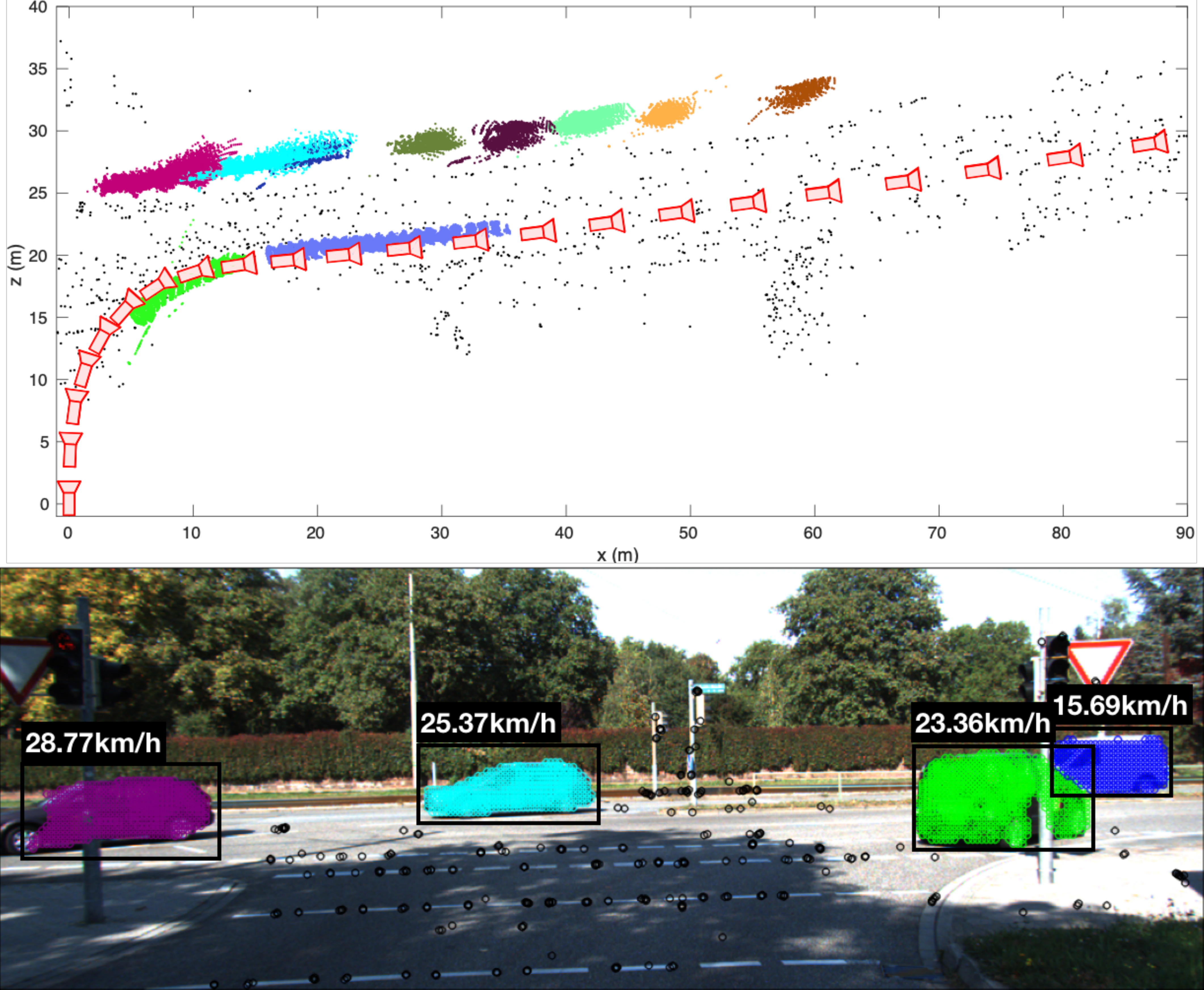}
 \caption{\textbf{Results of our VDO-SLAM system.} (Top) A full map including camera trajectory in red, static background points in black and points on moving objects colour coded by their instance. (Bottom) Detected 3D points on the static background and the objects' body, and the estimated object speed. Black circles represents static points, and each object is shown with a different colour.}
 \label{fig:showcase}
\end{figure}
%
% -- A definition of dynamic SLAM
\IEEEPARstart{T}he ability of a robot to build a model of the environment, often called map, and to localise itself within this map is a key factor in enabling autonomous robots to operate in real world environments. Creating these maps is achieved by fusing multiple sensor measurements into a consistent representation using estimation techniques such as Simultaneous Localisation And Mapping (SLAM). SLAM is a mature research topic and have already revolutionised a wide range of applications from mobile robotics, inspection, entertainment and film production to exploration and monitoring of natural environments, amongst many others. However, most of the existing solutions to SLAM rely heavily on the assumption that the environment is predominantly static. 

%comprises simultaneously estimating a robot's state and building a model representation of its environment. 
%It has long been accepted to assume that the world is static when solving the SLAM problem, and to estimate a robot state and a map of the static structure of the environment. 
%It is, however, undeniable that the real world is highly dynamic. 
%While many accurate and efficient solutions to the problem exist, current SLAM algorithms can be easily induced to fail in highly dynamic environments~\cite{Cadena16tro}. 
The conventional techniques to deal with dynamics in SLAM is to either treat any sensor data associated with moving objects as outliers and remove them from the estimation process (\cite{Hahnel02iros,Hahnel03icra,Wolf05autonrobot,Zhao08icra,bescos18ral}), or detect moving objects and track them separately using traditional multi-target tracking approaches (\cite{Wang03icra,Miller07icra,Rogers10iros,Kundu11iccv}). 
The former technique excludes information about dynamic objects in the scene, and generates static only maps. 
The accuracy of the latter depends on the camera pose estimation, which is more susceptible to failure in complex dynamic environments.
Increased presence of autonomous systems in dynamic environments is driving the community to challenge the static world assumption that underpins most existing open-source SLAM algorithms. 
In this paper, we redefine the term ``mapping'' in SLAM to be concerned with a \emph{spatiotemporal representation of the world}, as opposed to the concept of a static map that has long been the emphasis of the classical SLAM algorithms. %including SLAM systems that can operate in dynamic environments by excluding the dynamics of the world. 
Our approach focuses on accurately estimate the motion of all dynamic entities in the environment including the robot and other moving objects in the scene, this information being highly relevant in the context of robot path planning and navigation in dynamic environments.

%It is far more interesting for dynamic SLAM algorithms to estimate the motion of dynamic objects in the environment, rather than their current state $-$not where they are, but how they move. 

% -- Object motion estimation from 2D to 3D to SE(3) TODO_MOVE FROM HERE!!!!!!
Existing scene motion estimation techniques mainly rely on optical flow estimation (\cite{yamaguchi14eccv,sun2010cvpr,Sun18cvpr,ilg2017cvpr}) and scene flow estimation (\cite{vogel2013iccv,menze2015cvpr,liu2019cvpr,jiang2019iccv}). Optical flow records the scene motion by estimating the velocities associated with the movement of brightness patterns on an image plane. %As it is characterised on the image plane, optical flow involves both camera and scene motion, which may fail in degenerate cases where the scene motion is close to the camera motion. 
Scene flow, on the other hand, describes the 3D motion field of a scene observed at different instants of time.
Those techniques only estimate linear translation of individual pixels or 3D points in the scene, and are not exploiting the collective behaviour points on rigid objects failing to describe the full $\SE$ motion of objects in the scene. In this paper we explore this collective behaviour of points on individual objects to obtain accurate and robust motion estimation of the objects in the scene while simultaneously localising the robot and map the environment. 

A typical SLAM system consists of a front-end module, that processes the raw data from the sensors and a back-end module, that integrates the obtained information (raw and higher-level information) into a probabilistic estimation framework. 
Simple primitives such as 3D locations of salient features are commonly used to represent the environment. 
This is largely a consequence of the fact that points are easy to detect, track and integrate within the SLAM estimation problem.

Feature tracking has been more reliable and robust with the advances in deep learning to provide algorithms that can reliably estimate the 2$D$ optical flow associated with the apparent motion of every pixel on an image in a dense manner. 
A task that is particularly important for data association and that has been otherwise challenging in dynamic environments using classical feature tracking methods. \\
Other primitives such as lines and planes (\cite{De14autonrobot,Kaess15icra,Henein17iros,Hsiao17icra}) or even objects (\cite{Mu16iros,Salas13cvpr,Yang19tro}) have been considered in order to provide richer map representations. 
To incorporate such information in existing geometric SLAM algorithms, either a dataset of 3$D$-models of every object in the scene must be available a priori (\cite{Salas13cvpr,Galvez16ras}) or the front end must explicitly provide object pose information in addition to detection and segmentation (\cite{MOT16,Byravan17icra,Wohlhart15cvpr}) adding a layer of complexity to the problem. 
The requirement for accurate 3$D$-models severely limits the potential domains of application, while to the best of our knowledge, multiple object tracking and 3$D$ pose estimation remain a challenge to learning techniques. 
There is a clear need for an algorithm that can exploit the powerful detection and segmentation capabilities of modern deep learning algorithms (\cite{he18pami,Bolya19iccv}) without relying on additional pose estimation or object model priors, an algorithm that operates at feature-level with the awareness of an object concept. 

%% -- Contribution
While the problems of SLAM and object motion tracking/estimation are long studied in isolation in the literature, recent approaches try to solve the two problems in a unified framework (\cite{Henein20icra,Huang2019iccv}). 
However, they both focus on the SLAM back-end instead of a full system, resulting in a severely limited performance in real world scenarios.
In this paper, we carefully integrate our previous works (\cite{zhang20iros,Henein20icra}) and propose VDO-SLAM, a novel feature-based stereo/RGB-D dynamic SLAM system, that leverages image-based semantic information to simultaneously localise the robot, map the static and dynamic structure, and track motions of rigid objects in the scene. 
Different to~\cite{Henein20icra}, we rely on a denser object feature representation to ensure robust tracking, and propose new factors to smoothen the motion of rigid objects in urban driving scenarios. 
Different to~\cite{zhang20iros}, an improved robust feature and object tracking method is proposed, with the ability to handle indirect occlusions resulting from the failure of semantic object segmentation. 
In summary, the contributions of this work are:
\begin{itemize}
	\setlength\itemsep{0em}
	\item a novel formulation to model dynamic scenes in a unified estimation framework over robot poses, static and dynamic $3$D points, and object motions.

	\item accurate estimation for $\SE$ motion of dynamic objects that outperforms state-of-the-art algorithms, as well as a way to extract objects' velocity in the scene,

	\item a robust method for tracking moving objects exploiting semantic information with the ability to handle indirect occlusions resulting from the failure of semantic object segmentation,

	\item a demonstrable \emph{full system} in complex and compelling real-world scenarios.
\end{itemize}
To the best of our knowledge, this is the first full dynamic SLAM system that is able to achieve motion segmentation, dynamic object tracking, and estimate the camera poses along with the static and dynamic structure, the full $\SE$ pose change of every rigid object in the scene, extract velocity information, and be demonstrable in real-world outdoor scenarios (see Fig.~\ref{fig:showcase}). 
We demonstrate the performance of our algorithm on real datasets and show capability of the proposed system to resolve rigid object motion estimation and yield motion results that are comparable to the camera pose estimation in accuracy and that outperform state-of-the-art algorithms by an order of magnitude in urban driving scenarios.

%% -- Roadmap
The remainder of this paper is structured as follows, in the following Section~\ref{sec:related} we discuss the related work.  In Section~\ref{sec:method} and~\ref{sec:system} we describe the proposed algorithm and system. 
We introduce the experimental setup, followed by the results and evaluations in Section~\ref{sec:expe}. 
We summarise and offer concluding remarks in Section~\ref{sec:conclusion}.

\section{Related Work}
\label{sec:related}

In the past two decades, the study of SLAM for dynamic environments has become more and more popular in the community, with a considerable amount of algorithms being proposed to solve the dynamic SLAM problem. 
Motivated by different goals to achieve, solutions in the literature can be mainly divided into three categories. 

The first category aims to explore robust SLAM against dynamic environments. 
Early methods in this category (\cite{Hahnel03icra,alcantarilla12icra,tan13ismar}) normally detect and remove the information drawn from dynamic foreground, which is seen as degrading the SLAM performance. 
More recent methods on this track tend to go further by not just removing the dynamic foreground, but also inpainting or reconstructing the static background that is occluded by moving targets. 
\cite{bescos18ral} present dynaSLAM that combines classic geometry and deep learning-based models to detect and remove dynamic objects, then inpaint the occluded background with multi-view information of the scene. 
Similarly, a Light Field SLAM front-end is proposed by~\cite{kaveti2020arxiv} to reconstruct the occluded static scene via Synthetic Aperture Imaging (SAI) technics. 
Different from~\cite{bescos18ral}, features on the reconstructed static background are also tracked and used to achieve better SLAM performance. 
The above state-of-the-art solutions achieve robust and accurate estimation by discarding the dynamic information. 
However, we argue that this information has potential benefits for SLAM if it is properly modelled. 
Furthermore, understanding dynamic scenes in addition to SLAM is crucial for many other robotics tasks such as planning, control and obstacle avoidance, to name a few. 

Approaches of the second category performs SLAM and Moving Objects Tracking (MOT) separately, as an extension to conventional SLAM for dynamic scene understanding (\cite{Wang07ijrr,Kundu11iccv,Reddy15iros,Barsan2018icra}). 
\cite{Wang07ijrr} developed a theory for performing SLAM with Moving Objects Tracking (SLAMMOT). 
In the latest version of their SLAM with detection and tracking of moving objects, the estimation problem is decomposed into two separate estimators (moving and stationary objects) to make it feasible to update both filters in real time. 
\cite{Kundu11iccv} tackle the SLAM problem with dynamic objects by solving the problems of Structure from Motion (SfM) and tracking of moving objects in parallel, and unifying the output of the system into a 3D dynamic map containing the static structure and the trajectories of moving objects. 
Later in~\cite{Reddy15iros}, the authors propose to integrate semantic constraints to further improve the 3D reconstruction. 
The more recent work \cite{Barsan2018icra} present a stereo-based dense mapping algorithm in a SLAM framework, with the advantage of accurately and efficiently reconstructing both static background and moving objects in large scale dynamic environments. 
The listed algorithms above have proven that combining multiple objects tracking with SLAM is doable and applicable for dynamic scene exploration. 
To take a step further by proper exploiting and establishing the spatial and temporal relationships between the robot, static background, stationary and dynamic objects, we show in this paper that the problems of SLAM and multi-object tracking are mutually beneficial. 

The last and most active category is object SLAM, which usually includes both static and dynamic objects. 
Algorithms in this class normally require specific modelling and representation of 3D object, such as 3D shape (\cite{salas2013cvpr,tateno2016icra,sucar2020arXiv}), surfel~\cite{runz2018ismar} or volumetric~\cite{xu2019icra} model, geometric model such as ellipsoid (\cite{hosseinzadeh2019icra,nicholson2018ral}) or 3D bounding box (\cite{Yang19tro,li2018eccv,li2020cvpr,bescos2021ral}), etc., to extract high-level primitive (e.g., object pose) and integrate into a SLAM framework. 
\cite{salas2013cvpr} is one of the earliest works to introduce an object-oriented SLAM paradigm, which represents cluttered scene in object level and constructs an explicit graph between camera and object poses to achieve joint pose-graph optimisation. 
Later, \cite{tateno2016icra} propose a novel 3D object recognition algorithm to ensure the system robustness and improve the accuracy of estimated object pose. 
The high-level scene representation enables real-time 3D recognition and significant compression of map storage for SLAM. 
Nevertheless, a database of pre-scanned or pre-trained object models has to be created in advance. 
To avoid prebuilt database, representing objects using surfel or voxel element in a dense manner starts to gain popularity, along with RGB-D cameras becoming widely used. 
\cite{runz2018ismar} present MaskFusion that adopts surfel representation to model, track and reconstruct objects in the scene, while \cite{xu2019icra} apply an octree-based volumetric model to objects and build multi-object dynamic SLAM system. 
Both methods succeed to exploit object information in a dense RGB-D SLAM framework, without prior knowledge of object model. 
Their main interest, however, is the 3D object segmentation and consistent fusion of the dense map rather than the estimation of the motion of the objects. 

Lately, the use of basic geometric models to represent objects becomes a popular solution due to the less complexity and easy integration into a SLAM framework. 
In QuadricSLAM~\cite{nicholson2018ral}, detected objects are represented as ellipsoids to compactly parametrise the size and 3D pose of an object. 
In this way, the quadric parameters are directly constrained as geometric error and formulated together with camera poses in a factor graph SLAM for joint estimation. 
\cite{Yang19tro} propose to combine 2D and 3D object detection with SLAM for both static and dynamic environments. 
Objects are represented as high-quality cuboids and optimized together with points and cameras through multi-view bundle adjustment. 
While both methods prove the mutual benefit between detected object and SLAM, their main focus is on object detection and SLAM primarily for static scenarios. 
In this paper, we take this direction further to tackle the challenging problem of dynamic object tracking within a SLAM framework, and exploit the relationships between moving objects and agent robot, static and dynamic structures for potential advantages.

Apart from the dynamic SLAM categories, the literature of 6-DoF object motion estimation is also crucial for dynamic SLAM problem. 
Quite a few methods have been proposed in the literature to estimate $\SE$ motion of objects in a visual odometry or SLAM framework (\cite{dewan2016icra,Judd18iros,huang2020cvpr}). 
\cite{dewan2016icra} present a model-free method for detecting and tracking moving objects in 3D LiDAR scans. 
The method sequentially estimates motion models using RANSAC~\cite{fischler1981cacm}, then segments and tracks multiple objects based on the models by a proposed Bayesian approach. 
In~\cite{Judd18iros}, the authors address the problem of simultaneous estimation of ego and third-party $\SE$ motions in complex dynamic scenes using cameras. 
They apply multi-model fitting techniques into a visual odometry pipeline and estimate all rigid motions within a scene. 
In later work, \cite{huang2020cvpr} present ClusterVO that is able to perform online processing for multiple motion estimations. 
To achieve this, a multi-level probabilistic association mechanism is proposed to efficiently track features and detections, then a heterogeneous Conditional Random
Field (CRF) clustering approach is applied to jointly infer cluster segmentations, with a sliding-window optimization for clusters in the end. 
While the above proposed methods represent an important step forward to the Multi-motion Visual Odometry (MVO) task, the study of spacial and temporal relationships is not fully explored but is arguably important. 
Therefore, by carefully considering the pros and cons in the literature of SLAM+MOT, object SLAM and MVO, this paper proposes a visual dynamic object-aware SLAM system that is able to achieve robust ego and object motion tracking, as well as consistent static and dynamic mapping in a novel SLAM formulation.

%%%%%%%%%%%%%%%%%%%%%%%%%%%%%%%%%%%%%%%%%%%%%%%%%%%%%%%%%%%%%%%%%%%%%%%%%%%%%%%%%%%%%%%%%%%%%%%%%%%%%%%%%%%%%%%

\section{Methodology}
\label{sec:method}

Before discussing details of the proposed system pipeline, as shown in Fig.~\ref{fig:system_overview}, this section covers the mathematical details of the core components in the system. 
Variables and notations are first introduced, including the novel way of modelling the motion of a rigid-object in a model free manner. 
Then we show how the camera pose and object motion are estimated in the tracking component of the system. 
Finally, a factor graph optimisation is proposed and applied in the mapping component, to refine the camera poses and object motions, and build a global consistent map including static and dynamic structure. 

% In this section we show how to model the motion of a rigid-object in a model free manner based on point tracking.
% We propose a factor graph optimisation to estimate the camera and object motion.

% In the tracking component of our system, shown in Fig.~\ref{fig:system_overview}, the cost function chosen to estimate the camera pose and object motion (described in Section~\ref{sec:mot_estimate}) is associated with the $3$D-$2$D re-projection error and is defined on the image plane.
% Since the noise is better characterised in image plane, this yields more accurate results for camera localisation (\cite{Nister04cvpr}).
% Moreover, based on this error term, we propose a novel formulation to jointly optimise the optical flow along with the camera pose and the object motion, to ensure a robust tracking of points (described in Section ~\ref{sec:FlowRefine}).
% In the mapping module, a $3$D error cost function is used to ensure best results of $3$D structure and object motions estimation as described in Section~\ref{sec:graph_opt}.

\begin{figure*}[ht]
	\centering
	\includegraphics[width=0.85\linewidth,trim=0mm 118mm 0mm 45mm,clip]{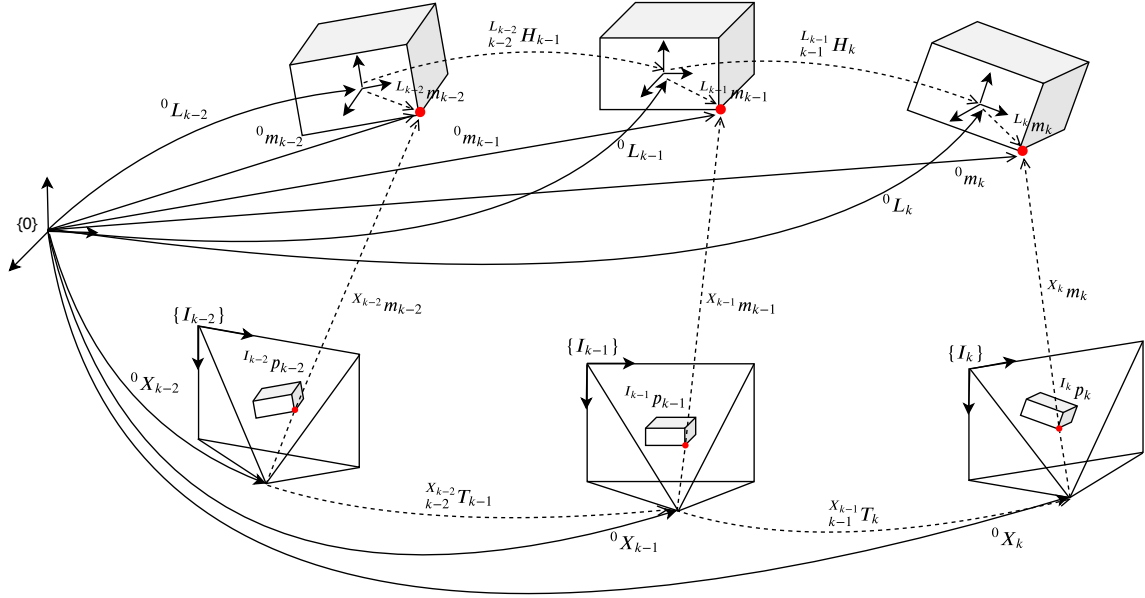}
	\caption{\textbf{Notation and coordinate frames.} Solid curves represent camera and object poses in inertial frame; $^{0}{\mathbf{X}}$ and $^{0}{\mathbf{L}}$ respectively, and dashed curves their respective motions in body-fixed frame. Solid lines represent $3$D points in inertial frame, and dashed lines represent $3$D points in camera frames.}
	\label{fig:coordinate_frames}
\end{figure*}

\subsection{Background and Notation}

\subsubsection{Coordinate Frames}
Let $^{0}\mathbf{X}_{k}, ^{0}\mathbf{L}_{k} \in \SE$ be the robot/camera and the object $3$D pose respectively, at time ${k}$ in a global reference frame $0$, with $k \in \mathcal{T}$ the set of time steps. Note that calligraphic capital letters are used in our notation to represent sets of indices.
% The set of all camera poses is defined as: \mbox{$\boldsymbol{\Theta}_{X} = \{{}^{0}\mathbf{X}_{k}\ |\ k \in \mathcal{T}\}$}.
Fig.~\ref{fig:coordinate_frames} shows these pose transformations as solid curves.

% We deliberately choose not to include the object $3$D pose as a random variable in the estimation, and however express the rigid object pose change in terms of the points that reside on the object as shown in the following.
% For this purpose, we define $\mathbf{I}_k$ the reference frame associated with the image captured by the camera at time $k$ chosen at the top left corner of the image.

\subsubsection{Points}
Let $^{0}\mathbf{m}_{k}^{i}$ be the homogeneous coordinates of the $i^{th}$ $3$D point at time ${k}$, with \mbox{$^{0}\mathbf{m}^{i} = \left [ {m}_{x}^{i},  {m}_{y}^{i}, {m}_{z}^{i}, 1 \right ]\tr \in \E^3$} and $\ i \in \mathcal{M}$ the set of points.
% We define the set of all $3$D points as: $\boldsymbol{\theta}_{M} = \{^{0}\mathbf{m}_{k}^{i}\ | \ i \in \mathcal{M}, k \in \mathcal{T}\}$ with $\mathcal{M}$ the set of all points.
We write a point in robot/camera frame as \mbox{$^{{X}_{k}}{\mathbf{m}}_{k}^{i} = ^{0}\mathbf{X}_{k}\inv \: {}^{0}{\mathbf{m}}_{k}^{i}$}.
% Let \mbox{$^{{I}_{k}}\mathbf{P}_{k} = \{^{{I}_{k}}\mathbf{p}^{i}_{k}\ | \ i \in \mathcal{M}, k \in \mathcal{T}\}$} be a set of projected points onto the image frame ${\mathbf{I}_{k}}$, with \mbox{$^{{I}_{k}}\mathbf{p}^{i}_{k} = \left [ {u}^{i}, {v}^{i}, 1 \right ] \in \E^2$} representing the image pixel location in homogeneous coordinates.
% The pixel location on frame $\mathbf{I}_{k}$ corresponding to the homogeneous $3$D point $^{\mathbf{X}_{k}}\mathbf{m}^{i}$ is obtained via the projection function $\pi(\cdot)$ as follows:

Define $\mathbf{I}_k$ the reference frame associated with the image captured by the camera at time $k$ chosen at the top left corner of the image, and let \mbox{$^{{I}_{k}}\mathbf{p}^{i}_{k} = \left [ {u}^{i}, {v}^{i}, 1 \right ] \in \E^2$} be the pixel location on frame $\mathbf{I}_{k}$ corresponding to the homogeneous $3$D point $^{\mathbf{X}_{k}}\mathbf{m}^{i}_{k}$, which is obtained via the projection function $\pi(\cdot)$ as follows:
\begin{eqnarray}
\label{eq:proj}
{}^{{I}_{k}}\mathbf{p}^{i}_{k}
= \pi(^{{X}_{k}}\mathbf{m}^{i}_{k}) = \mathbf{K} \: ^{{X}_{k}}\mathbf{m}^{i}_{k}\:,
\end{eqnarray}
where $\mathbf{K}$ is the camera intrinsics matrix.

% The camera and/or object motions both produce an optical flow \mbox{${}^{{I}_{k}}\boldsymbol{\Phi} = \{{}^{{I}_{k}}\boldsymbol{\phi}^{i}\ | \ i \in \mathcal{M}\}$}, where \mbox{${}^{{I}_{k}}\boldsymbol{\phi}^{i} \in \R^2$} is the corresponding optical flow of pixel \mbox{${}^{{I}_{k-1}}\mathbf{p}^{i}_{k-1}$} and its correspondence \mbox{${}^{{I}_{k}}\tilde{\mathbf{p}}^{i}_{k}$} in frame $k$, that is given by:
The camera and/or object motions both produce an optical flow \mbox{${}^{{I}_{k}}\boldsymbol{\phi}^{i} \in \R^2$} that is the displacement vector indicating the motion of pixel \mbox{${}^{{I}_{k-1}}\mathbf{p}^{i}_{k-1}$} from image frame ${I}_{k-1}$ to ${I}_{k}$, and is given by:
\begin{eqnarray}
\label{eq:of}
{}^{{I}_{k}}\boldsymbol{\phi}^{i} = {}^{{I}_k}\tilde{\mathbf{p}}^{i}_{k} - {}^{{I}_{k-1}}\mathbf{p}^{i}_{k-1}\:.
\end{eqnarray}
Here ${}^{{I}_k}\tilde{\mathbf{p}}^{i}_{k}$ is the correspondence of ${}^{{I}_{k-1}}\mathbf{p}^{i}_{k-1}$ in ${I}_{k}$. 
Note that, we overload the same notation to represent the 2D pixel coordinates $\in \R^2$.
In this work, we leverage optical flow to find correspondences between consecutive frames.
% Here we overload the notation and also use ${}^{{I}_{k-1}}{\mathbf{p}}^{i}_{k-1}$ and ${}^{{I}_k}\tilde{\mathbf{p}}^{i}_{k}$ to represent the 2D pixel coordinates $\in \R^2$. In this work, we leverage optical flow to find correspondences between consecutive frames.

\subsubsection{Object and 3D Point Motions}
The object motion between times $k-1$ and $k$ is described by the homogeneous transformation \mbox{$\prescript{L_{k-1}}{k-1}{\mathbf{H}}^{}_{k} \in \SE$} according to:
\begin{eqnarray}
\label{eq:obj_motion_bff}
\prescript{L_{k-1}}{k-1}{\mathbf{H}}^{}_{k} = ^{0}\mathbf{L}_{k-1}\inv \: ^{0}\mathbf{L}_{k}\:.
\end{eqnarray}
Fig.~\ref{fig:coordinate_frames} shows these motion transformations as dashed curves.
We write a point in its corresponding object frame as \mbox{$^{L_k}{\mathbf{m}}_{k}^{i} = {}^{0}\mathbf{L}_{k}\inv \: {}^{0}{\mathbf{m}}_{k}^{i}$} (shown as a dashed vector from the object reference frame to the red dot in Fig.~\ref{fig:coordinate_frames}), substituting the object pose at time $k$ from~\eqref{eq:obj_motion_bff}, this becomes:
\begin{eqnarray}
\label{eq:point_motion_nonrigid}
^{0}{\mathbf{m}}_{k}^{i} = {}^{0}\mathbf{L}_{k}\: ^{L_k}{\mathbf{m}}_{k}^{i}  = {}^{0}\mathbf{L}_{k-1} \: \prescript{L_{k-1}}{k-1}{\mathbf{H}}^{}_{k} \: ^{L_k}{\mathbf{m}}_{k}^{i}\:.
\end{eqnarray}
Note that for rigid body objects, \mbox{$^{L_k}{\mathbf{m}}_{k}^{i}$} stays constant at \mbox{$^{L}{\mathbf{m}}^{i}$}, and \mbox{$^{L}{\mathbf{m}}^{i} = {}^{0}\mathbf{L}_{k}\inv \: {}^{0}{\mathbf{m}}_{k}^{i} = {}^{0}\mathbf{L}_{k+n}\inv \: {}^{0}{\mathbf{m}}_{k+n}^{i}$} for any integer $n \in \mathbb{Z}$.
Then, for rigid objects with \mbox{$n = -1$}, \eqref{eq:point_motion_nonrigid} becomes:
\begin{eqnarray}
\label{eq:point_motion_rigid}
^{0}{\mathbf{m}}_{k}^{i} = {}^{0}\mathbf{L}_{k-1} \:
\prescript{L_{k-1}}{k-1}{\mathbf{H}}^{}_{k} \:  {}^{0}\mathbf{L}_{k-1}\inv \: {}^{0}{\mathbf{m}}_{k-1}^{i}\:.
\end{eqnarray}
\eqref{eq:point_motion_rigid} is crucially important as it relates the same $3$D point on a rigid object in motion at consecutive time steps by a homogeneous transformation \mbox{$\prescript{0}{k-1}{\mathbf{H}}^{}_{k} := {}^{0}\mathbf{L}_{k-1} \: \prescript{L_{k-1}}{k-1}{\mathbf{H}}^{}_{k} \:  {}^{0}\mathbf{L}_{k-1}\inv$}.
This equation represents a \emph{frame change of a pose transformation}~\cite{Chirikjian17idetc}, and shows how the body-fixed frame pose change $\prescript{L_{k-1}}{k-1}{\mathbf{H}}^{}_{k}$ relates to the global reference frame pose change $\prescript{0}{k-1}{\mathbf{H}}^{}_{k}$.
The point motion in global reference frame is then expressed as:
\begin{eqnarray}
\label{eq:point_motion}
^{0}{\mathbf{m}}_{k}^{i} = \prescript{0}{k-1}{\mathbf{H}}^{}_{k} \: {}^{0}{\mathbf{m}}_{k-1}^{i}\:.
\end{eqnarray}
Equation~\eqref{eq:point_motion} is at the core of our motion estimation approach, as it
expresses the rigid object pose change in terms of the points that reside on the object in a model-free manner without the need to include the object $3$D pose as a random variable in the estimation.
%We deliberately choose not to include the object $3$D pose as a random variable in the estimation, and however express the rigid object pose change in terms of the points that reside on the object. 
Section~\ref{sec:objectMotion} details how this rigid object pose change is estimated based on the above equation.
% We define the set of all rigid motions: \mbox{$\boldsymbol{\Theta}_{H} = \{\prescript{0}{k-1}{\mathbf{H}}^{l}_{k}\ |\ k \in \mathcal{T}, l \in \mathcal{L}\}$}, with $\mathcal{L}$ the set of all object labels.
Here $\prescript{0}{k-1}{\mathbf{H}}_k \in \SE$ represents the object point motion in global reference frame; for the remainder of this document, we refer to this quantity as the object pose change or the object motion for ease of reading.
\subsection{Camera Pose and Object Motion Estimation}
\label{sec:mot_estimate}
The cost function chosen to estimate the camera pose and object motion is associated with the $3$D-$2$D re-projection error and is defined on the image plane.
Since the noise is better characterised in image plane, this yields more accurate results for camera localisation~\cite{Nister04cvpr}.
Moreover, based on this error term, we propose a novel formulation to jointly optimise the optical flow along with the camera pose and the object motion, to ensure a robust tracking of points.
In the mapping module, a $3$D error cost function is used in global optimization to ensure best results of $3$D structure and object motions estimation as later described in Section~\ref{sec:graph_opt}.

\subsubsection{Camera Pose Estimation}
Given a set of static $3$D points \mbox{\{${}^{0}\mathbf{m}_{k-1}^{i}\ |\ i \in \mathcal{M}, k \in \mathcal{T}\}$} observed at time $k-1$ in global reference frame, and the set of $2D$ correspondences \mbox{\{${}^{I_k}\tilde{\mathbf{p}}_{k}^{i}\ |\ i \in \mathcal{M}, k \in \mathcal{T}$\}} in image $\mathbf{I}_{k}$, the camera pose {$\prescript{0}{}{\mathbf{X}}_{k}$} is estimated via minimizing the re-projection error: %between these two sets.
%The error term for the observation ${}^{I_k}\tilde{\mathbf{p}}_{k}^{i}$ in image $\mathbf{I}_{k}$ is formulated as:
%
\begin{eqnarray}
\label{eq:proj_static_image}
% \mathbf{e}_{i}({}^{0}\mathbf{X}_{k}) =  {}^{I_k}\tilde{\mathbf{p}}_{k}^{i}-\pi({}^{0}\mathbf{X}^{-1}_{k}\:{}^{0}\mathbf{m}_{k-1}^{i})\:.
\mathbf{e}_{i}(^{0}\mathbf{X}_{k}) =  {}^{I_k}\tilde{\mathbf{p}}_{k}^{i}-\pi({}^{0}\mathbf{X}^{-1}_{k}\:{}^{0}\mathbf{m}_{k-1}^{i})\:.
\end{eqnarray}
We parameterise the $\SE$ camera pose by elements of the Lie-algebra \mbox{$\mathbf{x}_k \in \se$}:  
%of $\SE$:
%
\begin{eqnarray}
{}^{0}\mathbf{X}_{k} = \exp({}^{0}{\mathbf{x}}_{k})\:,
\label{eq:X-se3}
\end{eqnarray}
and define ${}^{0}{\mathbf{x}}_{k}^\vee \in \R^6$ with the \emph{vee} operator a mapping from $\se$ to $\R^6$.
%where ${}^{0}{\mathbf{x}}_{k} \in \R^6$ and the wedge operator is the standard lift into $\se$.
Using the Lie-algebra parameterisation of $\SE$ with the substitution of~\eqref{eq:X-se3} into~\eqref{eq:proj_static_image}, the solution of the least squares cost is given by:
\begin{eqnarray}
\label{eq:camera_pose_2D_cost}
% ^{0}{\mathbf{x}}_{k}^{*} = \underset{^{0}{\mathbf{x}}_{k}}{\argmin} \sum^{n_{b}}_{i}{ \rho_{h}( \mathbf{e}_{i}\tr(\mathbf{x})\:\Sigma_{p}\inv\:\mathbf{e}_{i}(\mathbf{x}))}
{^{0}{\mathbf{x}}_{k}^{*\vee}} = \underset{^{0}{\mathbf{x}}_{k}^\vee}{\argmin} \sum_{i}^{n_b} \rho_h\left( \mathbf{e}_{i}^\top(^{0}{\mathbf{x}}_{k})\:\Sigma_{p}\inv\:\mathbf{e}_{i}(^{0}{\mathbf{x}}_{k}) \right)
\end{eqnarray}
for all $n_{b}$ visible $3$D-$2$D static background point correspondences between consecutive frames. Here $\rho_{h}$ is the Huber function~\cite{Huber92bs}, and $\Sigma_p$ is the covariance matrix associated with the re-projection error. 
The estimated camera pose is given by ${}^{0}{\bf{X}}_{k}^* = \exp({}^{0}{{\mathbf{x}}_{k}^{*}})$ and is found using the Levenberg-Marquardt algorithm to solve for \eqref{eq:camera_pose_2D_cost}.

\subsubsection{Object Motion Estimation}
\label{sec:objectMotion}
Analogous to the camera pose estimation, a cost function based on re-projection error is constructed to solve for the object motion \mbox{$\prescript{0}{k-1}{\mathbf{H}}^{}_{k}$}.
Using \eqref{eq:point_motion}, the error term between the re-projection of an object $3$D point and the corresponding $2D$ point in image $\mathbf{I}_{k}$ is:
\begin{eqnarray}\nonumber
\label{eq:proj_dynamic_image}
\mathbf{e}_{i}(\prescript{0}{k-1}{\mathbf{H}}^{}_{k}) := & {}^{I_k}\tilde{\mathbf{p}}_{k}^{i}-\pi({}^{0}\mathbf{X}^{-1}_{k} \: \prescript{0}{k-1}{\mathbf{H}}^{}_{k} \:{}^{0}\mathbf{m}_{k-1}^{i}) \\
= & {}^{I_k}\tilde{\mathbf{p}}_{k}^{i}-\pi(\prescript{0}{k-1}{\mathbf{G}}^{}_{k} \: {}^{0}\mathbf{m}_{k-1}^{i})
\:,
\end{eqnarray}
where \mbox{$\prescript{0}{k-1}{\mathbf{G}}^{}_{k} \in \SE$}. 
Parameterising \mbox{$\prescript{0}{k-1}{\mathbf{G}}^{}_{k} := \exp \left(\prescript{0}{k-1}{\mathbf{g}}_{k}\right)$} with \mbox{$\prescript{0}{k-1}{\mathbf{g}}_{k}$} $ \in \se$, 
%representation of \mbox{$\prescript{0}{k-1}{\mathbf{g}}_{k} \in \R^6$}, 
the optimal solution is found via minimising:
\begin{flalign}
\label{eq:obj_mot_2D_cost}
% \prescript{0}{k-1}{\mathbf{g}}_{k}^{*} = \underset{\prescript{0}{k-1}{\mathbf{g}}_{k}}{\argmin} \sum^{n_{d}}_{i}{ \rho_{h}( \mathbf{e}_{i}\tr(\mathbf{g})\:\Sigma_{p}\inv\:\mathbf{e}_{i}(\mathbf{g}))}
\prescript{0}{k-1}{\mathbf{g}}_{k}^{*\vee} = \underset{\prescript{0}{k-1}{\mathbf{g}}_{k}^\vee}{\argmin} \sum_{i}^{n_d} \rho_h\left( \mathbf{e}_{i}^\top(\prescript{0}{k-1}{\mathbf{g}}_{k})\:\Sigma_{p}\inv\:\mathbf{e}_{i}(\prescript{0}{k-1}{\mathbf{g}}_{k}) \right)
\end{flalign}
given all $n_{d}$ visible $3$D-$2$D dynamic point correspondences on an object between frames $k-1$ and $k$. The object motion, \mbox{$\prescript{0}{k-1}{\mathbf{H}}_{k} = {}^{0}\mathbf{X}_{k} \: \prescript{0}{k-1}{\mathbf{G}}_{k}$} can be recovered afterwards.

\subsubsection{Joint Estimation with Optical Flow}
\label{sec:FlowRefine}
The camera pose and object motion estimation both rely on good image correspondences.
Tracking of points on moving objects can be very challenging due to occlusions, large relative motions and large camera-object distances.
In order to ensure a robust tracking of points, we follow our earlier work~\cite{zhang20iros} to refine the estimation of the optical flow jointly with the motion estimation.
%the technique proposed in this paper aims at refining the estimation of the optical flow jointly with the motion estimation.

For camera pose estimation, the error term in~\eqref{eq:proj_static_image} is reformulated considering~\eqref{eq:of} as:
\begin{eqnarray}
\label{eq:proj_static_image_of}
\mathbf{e}_{i}({}^{0}\mathbf{X}_{k},^{I_{k}}\boldsymbol{\phi}) =  {}^{I_{k-1}}\mathbf{p}_{k-1}^{i} + {}^{{I}_{k}}\boldsymbol{\phi}^{i} -\pi({}^{0}\mathbf{X}^{-1}_{k}\:{}^{0}\mathbf{m}_{k-1}^{i})\:.
\end{eqnarray}
Applying the Lie-algebra parameterisation of $\SE$ element, the optimal solution is obtained via minimising the cost function:
\begin{flalign}
\label{eq:cam_pose_flow_2D_cost}
% {\{{}^{0}{\mathbf{x}}_{k}^{*},{}^{k}{\boldsymbol{\Phi}}_{k}^{*}\}} =
% & \underset{\{{}^{0}{\mathbf{x}}_{k},{}^{k}{\boldsymbol{\Phi}}_{k}\}} {\argmin} \sum^{n_{b}}_{i} \Big\{{\rho}_{h}\big( \mathbf{e}_{i}\tr(\mathbf{x},\boldsymbol{\phi})\:\Sigma_{p}\inv\:\mathbf{e}_{i}(\mathbf{x},\boldsymbol{\phi})\big) \nonumber \\
% & + {\rho}_{h}\big( \mathbf{e}_{i}\tr(\boldsymbol{\phi})\:\Sigma_{\phi}\inv\:\mathbf{e}_{i}(\boldsymbol{\phi})\big)\Big\}\:,
{\{{}^{0}{\mathbf{x}}_{k}^{*\vee},{}^{k}{\boldsymbol{\Phi}}_{k}^{*}\}} =
& \underset{\{{}^{0}{\mathbf{x}}_{k}^\vee,{}^{k}{\boldsymbol{\Phi}}_{k}\}} {\argmin} \sum^{n_{b}}_{i} \Big\{ {\rho}_{h}\big( \mathbf{e}_{i}\tr(^{I_{k}}\boldsymbol{\phi}^{i})\:\Sigma_{\phi}\inv\:\mathbf{e}_{i}(^{I_{k}}\boldsymbol{\phi}^{i})\big)\:+\: \nonumber \\
& {\rho}_{h}\big( \mathbf{e}_{i}\tr(^{0}{\mathbf{x}}_{k},^{I_{k}}\boldsymbol{\phi}^{i})\:\Sigma_{p}\inv\:\mathbf{e}_{i}(^{0}{\mathbf{x}}_{k},^{I_{k}}\boldsymbol{\phi}^{i})\big)\Big\}\:,
\end{flalign}
where \mbox{${\rho}_{h}( \mathbf{e}_{i}\tr(^{I_{k}}\boldsymbol{\phi}^i)\:\Sigma_{\phi}\inv\:\mathbf{e}_{i}(^{I_{k}}\boldsymbol{\phi}^i) )$} is the regularization term with
\begin{eqnarray}
\label{eq:flow_regular}
\mathbf{e}_{i}(^{I_{k}}\boldsymbol{\phi}^i) = {}^{{I}_{k}}\hat{\boldsymbol{\phi}}^{i} - {}^{{I}_{k}}\boldsymbol{\phi}^{i}\:.
\end{eqnarray}
Here \mbox{${}^{{I}_{k}}\hat{\boldsymbol{\Phi}}^{i} = \{{}^{{I}_{k}}\hat{\boldsymbol{\phi}}^{i}\ |\ i \in \mathcal{M}, k \in \mathcal{T}\}$} is the initial optic-flow obtained through classical or learning-based methods, and $\Sigma_{\phi}$ is the associated covariance matrix. Analogously, the cost function for object motion in~\eqref{eq:obj_mot_2D_cost} combining optical flow refinement is given by
\begin{flalign}
\label{eq:obj_mot_flow_2D_cost}
% {\{{}^{0}_{k-1}{\mathbf{g}}_{k}^{*},{}^{k}{\boldsymbol{\Phi}}_{k}^{*}\}} =
% & \underset{\{{}^{0}_{k-1}{\mathbf{g}}_{k},{}^{k}{\boldsymbol{\Phi}}_{k}\}} {\argmin} \sum^{n_{d}}_{i} \Big\{{\rho}_{h}\big( \mathbf{e}_{i}\tr(\mathbf{g},\boldsymbol{\phi})\:\Sigma_{p}\inv\:\mathbf{e}_{i}(\mathbf{g},\boldsymbol{\phi})\big) \nonumber \\
% & + {\rho}_{h}\big( \mathbf{e}_{i}\tr(\boldsymbol{\phi})\:\Sigma_{\phi}\inv\:\mathbf{e}_{i}(\boldsymbol{\phi})\big)\Big\}\:.
{\{{}^{0}_{k-1}{\mathbf{g}}_{k}^{*\vee},{}^{k}{\boldsymbol{\Phi}}_{k}^{*}\}} =
& \underset{\{{}^{0}_{k-1}{\mathbf{g}}_{k}^\vee,{}^{k}{\boldsymbol{\Phi}}_{k}\}} {\argmin} \sum^{n_{d}}_{i} \Big\{ {\rho}_{h}\big( \mathbf{e}_{i}\tr(^{I_{k}}\boldsymbol{\phi}^i)\:\Sigma_{\phi}\inv\:\mathbf{e}_{i}(^{I_{k}}\boldsymbol{\phi}^i)\big) \:+\: \nonumber \\
& {\rho}_{h}\big( \mathbf{e}_{i}\tr(\prescript{0}{k-1}{\mathbf{g}}_{k},^{I_{k}}\boldsymbol{\phi}^i)\:\Sigma_{p}\inv\:\mathbf{e}_{i}(\prescript{0}{k-1}{\mathbf{g}}_{k},^{I_{k}}\boldsymbol{\phi}^i)\big)\Big\}\:.
\end{flalign}

\subsection{Graph Optimisation}
\label{sec:graph_opt}
\begin{figure}[ht]
	\centering
	\includegraphics[width=.9\linewidth]{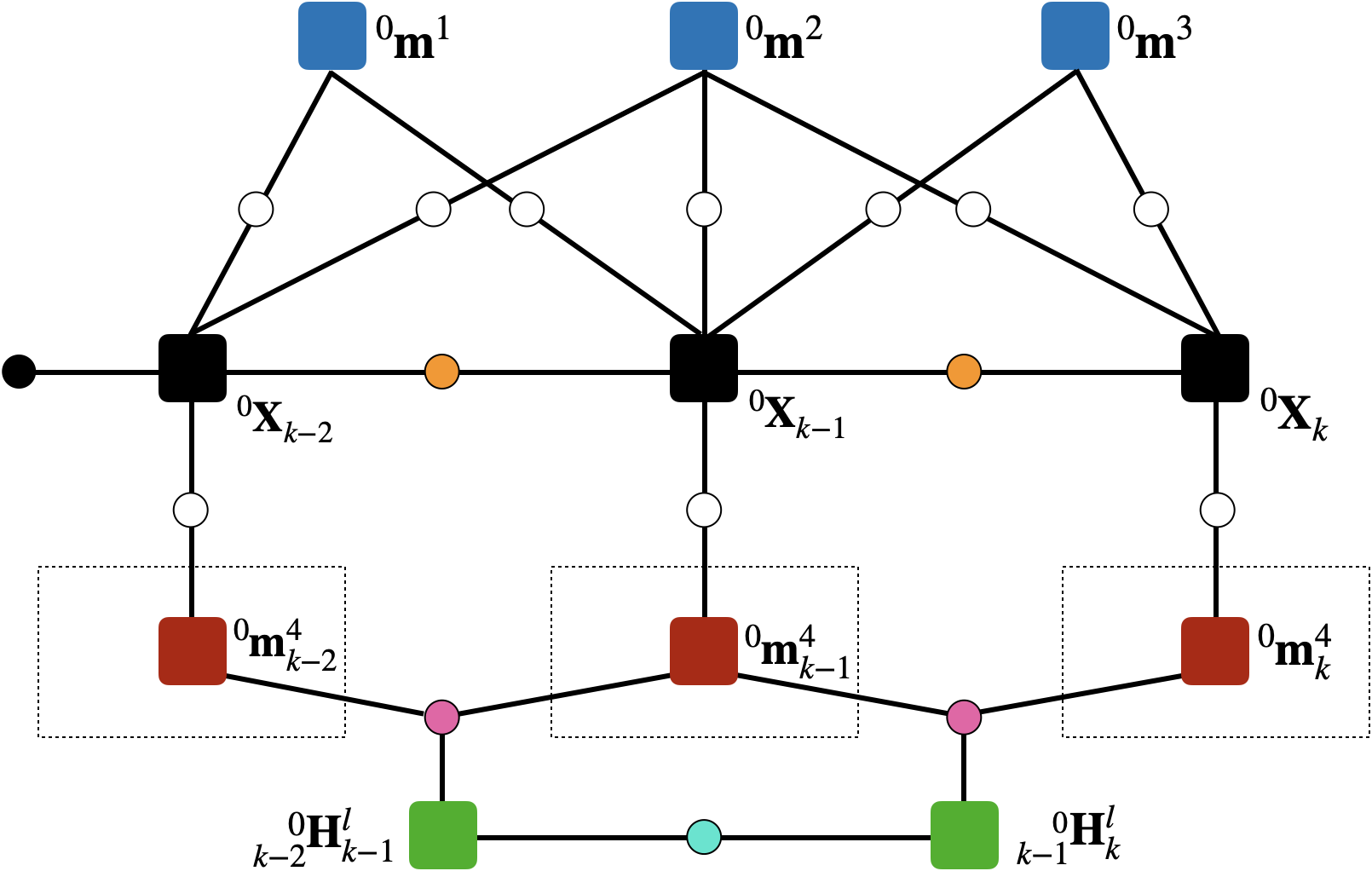}
	\caption{\textbf{Factor graph representation of an object-aware SLAM with a moving object.} Black squares stand for the camera poses at different time steps, blue for static points, red for the same dynamic point on an object (dashed box) at different time steps and green for the object pose change between time steps. For ease of visualisation, only one dynamic point is drawn here. A prior factor is shown as a black circle, odometry factors are shown as orange, point measurement factors as white and point motion factors as magenta. A smooth motion factor is shown as cyan circle.}
	%, however, at the time of estimation, all points on a detected dynamic object are used.
	\label{fig:factor-graph}
\end{figure}

% The proposed approach refines the camera poses, the static and dynamic structure as well as the motion of the dynamic rigid objects.
The proposed approach formulates the dynamic SLAM as a graph optimisation problem, to refine the camera poses and object motions, and build a global consistent map including static and dynamic structure.
We model the dynamic SLAM problem as a factor graph as the one demonstrated in Fig.~\ref{fig:factor-graph}.
The factor graph formulation is highly intuitive and has the advantage that it allows for efficient implementations of batch (\cite{Dellaert06ijrr,ceres-solver}) and incremental (\cite{Kaess11ijrr,Polok13rss,Ila17ijrr}) solvers.

Four types of measurements/observations are integrated into a joint optimisation problem; the $3$D point measurements, the visual odometry measurements, the motion of points on a dynamic object and the object smooth motion observations.

The $3$D point measurement model error \mbox{$\mathbf{e}_{i,k}(^{0}\mathbf{X}_k, ^{0}\mathbf{m}^i_k)$} is defined as:
\begin{eqnarray}
\mathbf{e}_{i,k}(^{0}\mathbf{X}_k, ^{0}\mathbf{m}^i_k) =
^{0}\mathbf{X}_k\inv \: ^{0}\mathbf{m}^i_k - \mathbf{z}^i_k \: .
\label{eq:3DPointMeasFactor}
\end{eqnarray}
Here \mbox{$\textbf{z} = \{\mathbf{z}_{k}^{i} \ | \ i \in \mathcal{M}, k \in \mathcal{T}\}$} is the set of all $3$D point measurements at all time steps, with cardinality $n_z$ and \mbox{$\mathbf{z}_{k}^{i} \in {\rm I\!R}^3$}. 
The $3$D point measurement factors are shown as white circles in Fig.~\ref{fig:factor-graph}.
\\
The tracking component of the system provides a high-quality ego-motion via 3D-2D error minimization, which can be used as an odometry measurement to constrain camera poses in the graph. 
The visual odometry model error $\mathbf{e}_{k}(^{0}\mathbf{X}_{k-1}, ^{0}\mathbf{X}_{k})$ is defined as:
\begin{eqnarray}
\mathbf{e}_{k}(^{0}\mathbf{X}_{k-1}, ^{0}\mathbf{X}_{k}) = (^{0}\mathbf{X}_{k-1}\inv \: ^{0}\mathbf{X}_{k})\inv \prescript{X_{k-1}}{k-1}{\mathbf{T}}_k\:,
\label{eq:visualOdomFactor}
\end{eqnarray}
where \mbox{$\textbf{T} = \{\prescript{X_{k-1}}{k-1}{\mathbf{T}}_k \ | \ k \in \mathcal{T}\}$} is the odometry measurement set with $\prescript{X_{k-1}}{k-1}{\mathbf{T}}_k \in \SE$ and cardinality $n_o$.
The odometric factors are shown as orange circles in Fig.~\ref{fig:factor-graph}. 
\\
The motion model error of points on dynamic objects $\mathbf{e}_{i,l,k}(^0\mathbf{m}_{k}^i, \prescript{0}{k-1}{\mathbf{H}}_{k}^l, ^0\mathbf{m}_{k-1}^i)$ is defined as:
\begin{eqnarray}
\mathbf{e}_{i,l,k}(^0\mathbf{m}_{k}^i, \prescript{0}{k-1}{\mathbf{H}}_{k}^l, ^0\mathbf{m}_{k-1}^i) = 
\: ^0\mathbf{m}_{k}^i - {\prescript{0}{k-1}{\mathbf{H}}_{k}^l} \: ^0\mathbf{m}_{k-1}^i \:.
\label{eq:MotionGeneral}
\end{eqnarray}
The motion of all points on a detected rigid object $l$ are characterised by the same pose transformation \mbox{$\prescript{0}{k-1}{\mathbf{H}}_{k}^l \in \SE$} given by \eqref{eq:point_motion} and the corresponding factor, shown as magenta circles in Fig.~\ref{fig:factor-graph}, is a ternary factor which we call the \emph{motion model of a point on a rigid body}. 
\\
It has been shown that incorporating prior knowledge about the motion of objects in the scene is highly valuable in dynamic SLAM (\cite{Wang07ijrr, Henein20icra}).
%In our previous work (\cite{Henein20icra}), we have shown how the use of a constant motion assumption of vehicles on highways improves the estimation.
Motivated by the camera frame rate and the physics laws governing the motion of relatively large objects (vehicles) and preventing their motions to change abruptly, we introduce smooth motion factors to minimise the change in consecutive object motions, with the error term defined as:
\begin{eqnarray}
\mathbf{e}_{l,k}(\prescript{0}{k-2}{\mathbf{H}}_{k-1}^l, \prescript{0}{k-1}{\mathbf{H}}_{k}^l) = {\prescript{0}{k-2}{\mathbf{H}}_{k-1}^l}\inv \: {\prescript{0}{k-1}{\mathbf{H}}_{k}^l}.
\label{eq:smoothMotionFactor}
\end{eqnarray}
%
%with $\Sigma_{s}$ the smooth motion covariance matrix, and $n_s$ the total number of smooth motion factors.
The object smooth motion factor \mbox{$\mathbf{e}_{l,k}(\prescript{0}{k-2}{\mathbf{H}}_{k-1}^l, \prescript{0}{k-1}{\mathbf{H}}_{k}^l)$} is used to minimise the change between the object motion at consecutive time steps and is shown as cyan circles in Fig.~\ref{fig:factor-graph}.

Let $\boldsymbol{\theta}_{M} = \{^{0}\mathbf{m}_{k}^{i}\ | \ i \in \mathcal{M}, k \in \mathcal{T}\}$ be the set of all 3D points, and $\boldsymbol{\theta}_{X} = \{{}^{0}\mathbf{x}_{k}^\vee \ |\ k \in \mathcal{T}\}$ as the set of all camera poses. %, where the \emph{vee}-operator is a mapping from $\se$ to a minimal vector representation in $\R^6$.
We parameterise the $\SE$ object motion $\prescript{0}{k-1}{\mathbf{H}}^{l}_{k}$ by elements \mbox{${\prescript{0}{k-1}{\mathbf{h}}^{l}_{k}} \in \se$} the Lie-algebra of $\SE$:
\begin{eqnarray}
\prescript{0}{k-1}{\mathbf{H}}^{l}_{k} = \exp({\prescript{0}{k-1}{\mathbf{h}}^{l}_{k}})\:,
\label{eq:H-se3}
\end{eqnarray}
and define $\boldsymbol{\theta}_{H} = \{{\prescript{0}{k-1}{\mathbf{h}}^{l}_{k}}^\vee \ |\ k \in \mathcal{T}, l \in \mathcal{L}\}$ as the set of all object motions, with ${\prescript{0}{k-1}{\mathbf{h}}^{l}_{k}}^\vee \in \R^6$ and $\mathcal{L}$ the set of all object labels. 
Given \mbox{${\boldsymbol{\theta}} = {\boldsymbol{\theta}_{X}} \cup {\boldsymbol{\theta}_{M}} \cup {\boldsymbol{\theta}_{H}}$} as all the nodes in the graph, with the Lie-algebra parameterisation of $\SE$ for $\mathbf{X}$ and $\mathbf{H}$ (substituting~\eqref{eq:X-se3} in~\eqref{eq:3DPointMeasFactor} and~\eqref{eq:visualOdomFactor}, and substituting~\eqref{eq:H-se3} in~\eqref{eq:MotionGeneral} and~\eqref{eq:smoothMotionFactor}), the solution of the least squares cost is given by: 
\begin{multline}
\boldsymbol \theta^* = \argmin_{\boldsymbol \theta} \Big\{
\sum^{n_z}_{i,k} \rho_{h} \big( \mathbf{e}_{i,k}\tr({}^{0}{\mathbf{x}}_{k}, ^{0}\mathbf{m}^i_k) \: \Sigma_{z}\inv \: \mathbf{e}_{i,k}({}^{0}{\mathbf{x}}_{k}, ^{0}\mathbf{m}^i_k) \big) \: \\  + 
\sum^{n_o}_{k} \rho_{h}\big(\log(\mathbf{e}_{k}({}^{0}{\mathbf{x}}_{k-1}, {}^{0}{\mathbf{x}}_{k}))\tr \: \Sigma_{o}\inv \: \log(\mathbf{e}_{k}({}^{0}{\mathbf{x}}_{k-1}, {}^{0}{\mathbf{x}}_{k}))\big) \: \\ \hspace{-2.4cm} + 
\sum^{n_g}_{i,l,k} \rho_{h}\big(\mathbf{e}_{i,l,k}\tr(^{0}\mathbf{m}^i_{k}, {{\prescript{0}{k-1}{\mathbf{h}}^{l}_{k}}}, ^{0}\mathbf{m}^i_{k-1}) \:
\Sigma_{g}\inv \\ \: \mathbf{e}_{i,l,k}(^{0}\mathbf{m}^i_{k}, {\prescript{0}{k-1}{\mathbf{h}}^{l}_{k}}, ^{0}\mathbf{m}^i_{k-1})\big)\: \\ 
\hspace{-2cm} +
\sum^{n_s}_{l,k} \rho_{h}\big(\log(\mathbf{e}_{l,k}({\prescript{0}{k-2}{\mathbf{h}}^{l}_{k-1}}, {\prescript{0}{k-1}{\mathbf{h}}^{l}_{k}}))\tr \:
\Sigma_{s}\inv \\ \: \log(\mathbf{e}_{l,k}({\prescript{0}{k-2}{\mathbf{h}}^{l}_{k-1}}, {\prescript{0}{k-1}{\mathbf{h}}^{l}_{k}}))\big) \Big\}\:,
\label{eq:NLSMC}
\end{multline}
where $\Sigma_{z}$ is the $3$D point measurement noise covariance matrix, $\Sigma_{o}$ is the odometry noise covariance matrix, $\Sigma_{g}$ is the motion noise covariance matrix with $n_g$ the total number of ternary object motion factors, and $\Sigma_{s}$ the smooth motion covariance matrix, with $n_s$ the total number of smooth motion factors.
The non-linear least squares problem in~\eqref{eq:NLSMC} is solved using Levenberg-Marquardt method.

%%%%%%%%%%%%%%%%%%%%%%%%%%%%%%%%%%%%%%%%%%%%%%%%%%%%%%%%%%%%%%%%%%%%%%%%%%%%%%%%%%%%%%%%%%%%%%%%%%%%%%%%%%%%%%%

\section{System}
\label{sec:system}

\begin{figure*}[ht]
 \centering
 \includegraphics[width=2.0\columnwidth]{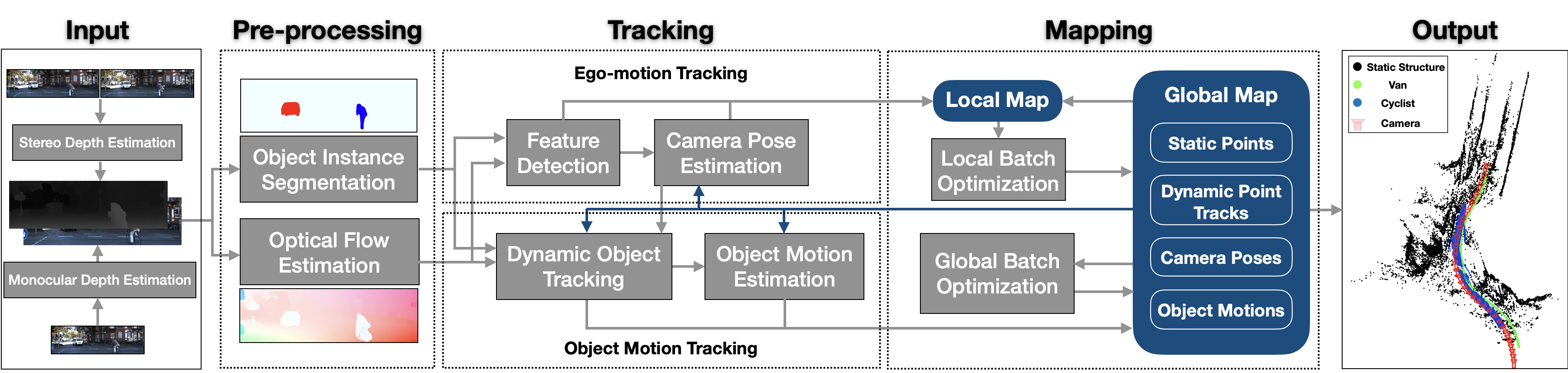}
 \caption{\textbf{Overview of our VDO-SLAM system.} Input images are first pre-processed to generate instance-level object segmentation and dense optical flow. These are then used to track features on static background structure and dynamic objects. Camera poses and object motions estimated from feature tracks are then refined in a global batch optimisation, and a local map is maintained and updated with every new frame. The system outputs camera poses, static structure, tracks of dynamic objects, and estimates of their pose changes over time.}
 \label{fig:system_overview}
\end{figure*}

In this section, we propose a novel object-aware dynamic SLAM system that robustly estimates both camera and object motions, along with the static and dynamic structure of the environment.
The full system overview is shown in Fig.~\ref{fig:system_overview}. 
The system consists of three main components: image pre-processing, tracking and mapping.

The input to the system is stereo or RGB-D images.
For stereo images, as a first step, we extract depth information by applying the stereo depth estimation method described in~\cite{yamaguchi14cvpr} to generate depth maps and the resulting data is treated as RGB-D.

Although this system was initially designed to be an RGB-D system, as an attempt to fully exploit image-based semantic information, we apply single image depth estimation to achieve depth information from monocular camera.
Our ``learning-based monocular'' system is monocular in the sense that only RGB images are used as input to the system, however the estimation problem is formulated using RGB-D data, where the depth is obtained using single image depth estimation.
%This learning-based monocular version of our system is used to compare our results vs CubeSLAM~\cite{Yang19tro}, a monocular state-of-the-art object SLAM system that performs well in dynamic environments.
%
\subsection{Pre-processing}
There are two challenging aspects that this module needs to fulfil.
First, to robustly separate static background and objects, and secondly to ensure long-term tracking of dynamic objects.
To achieve this, we leverage recent advances in computer vision techniques for instance level semantic segmentation and dense optical flow estimation in order to ensure efficient object motion segmentation and robust object tracking.
\subsubsection{Object Instance Segmentation}
Instance-level semantic segmentation is used to segment and identify potentially movable objects in the scene.
Semantic information constitutes an important prior in the process of separating static and moving object points, e.g., buildings and roads are always static, but cars can be static or dynamic.
Instance segmentation helps to further divide semantic foreground into different instance masks, which makes it easier to track each individual object.
Moreover, segmentation masks provide a ``precise'' boundary of the object body that ensures robust tracking of points on the object.
\subsubsection{Optical Flow Estimation}
The dense optical flow is used to maximise the number of tracked points on moving objects.
Most of the moving objects only occupy a small portion of the image. Therefore, using sparse feature matching does not guarantee robust nor long-term feature tracking.
Our approach makes use of dense optical flow to considerably increase the number of object points by sampling from all the points within the semantic mask.
Dense optical flow is also used to consistently track multiple objects by propagating a unique object identifier assigned to every point on an object mask.
Moreover, it allows to recover objects masks if semantic segmentation fails; a task that is extremely difficult to achieve using sparse feature matching.

\subsection{Tracking}
The tracking component includes two modules; the camera ego-motion tracking with sub-modules of feature detection and camera pose estimation, and the object motion tracking including sub-modules of dynamic object tracking and object motion estimation.%the camera ego-motion tracking and the object motion tracking.

\subsubsection{Feature Detection}
To achieve fast camera pose estimation, we detect a \emph{sparse} set of corner features and track them with optical flow.
At each frame, only inlier feature points that fit the estimated camera motion are saved into the map, and used to track correspondences in the next frame.
New features are detected and added, if the number of inlier tracks falls below a certain level ($1200$ in default).
These sparse features are detected on static background, i.e., image regions excluding the segmented objects.
\subsubsection{Camera Pose Estimation}
\label{sec:cam_pose}
The camera pose is computed using~\eqref{eq:cam_pose_flow_2D_cost} for all detected $3$D-$2$D static point correspondences.
To ensure robust estimation, a motion model generation method is applied for initialisation.
Specifically, the method generates two models and compares their inlier numbers based on re-projection error.
One model is generated by propagating the camera previous motion, while the other by computing a new motion transform using P$3$P~\cite{ke17cvpr} algorithm with RANSAC.
The motion model that generates most inliers is then selected for initialisation.
\subsubsection{Dynamic Object Tracking}
The process of object motion tracking consists of two steps.
In the first step, segmented objects are classified into static and dynamic.
Then we associate the dynamic objects across pairs of consecutive frames.
\\
$\bullet$
Instance-level object segmentation allows us to separate objects from background.
Although the algorithm is capable of estimating the motions of all the segmented objects, dynamic object identification helps reduce computational cost of the proposed system.
This is done based on scene flow estimation.
Specifically, after obtaining the camera pose $\prescript{0}{}{\mathbf{X}}_{k}$, the scene flow vector $\mathbf{f}^{i}_{k}$ describing the motion of a $3$D point ${}^{0}\mathbf{m}^{i}$ between frames $k-1$ and $k$, can be calculated as in~\cite{Lv18eccv}:
\begin{eqnarray}
\label{eq:sf}
\mathbf{f}^{i}_{k} = {}^{0}\mathbf{m}_{k-1}^{i}-{}^{0}\mathbf{m}_{k}^{i} = {}^{0}\mathbf{m}_{k-1}^{i} - ^{0}\mathbf{X}_{k} \: {}^{X_{k}}\mathbf{m}_{k}^{i}\:.
\end{eqnarray}
Unlike optical flow, scene flow$-$ideally only caused by scene motion$-$can directly decide whether some structure is moving or not.
Ideally, the magnitude of the scene flow vector should be zero for all static $3$D points.
However, noise or error in depth and matching complicates the situation in real scenarios.
To robustly handle this, we compute the scene flow magnitude of all the sampled points on each object. If the magnitude of the scene flow of a certain point is greater than a predefined threshold, the point is considered dynamic. This threshold was set to $0.12$ in all experiments carried in this work. An object is then recognised dynamic if the proportion of ``dynamic'' points is above a certain level ($30$\% of total number of points), otherwise static. Thresholds to identify if an object is dynamic were deliberately chosen as mentioned above, to be more conservative as the system is flexible to model a static object as dynamic and estimate a zero motion at every time step, however, the opposite would degrade the system's performance.
\\
$\bullet$
Instance-level object segmentation only provides single-image object labels.
Objects then need to be tracked across frames and their motion models propagated over time.
We propose to use optical flow to associate point labels across frames.
A point label is the same as the unique object identifier on which the point was sampled.
We maintain a finite tracking label set \mbox{$\mathcal{L}\subset \mathbb{N}$}, where \mbox{$l \in \mathcal{L}$} starts from $l=1$ for the first detected moving object in the scene.
The number of elements in $\mathcal{L}$ increases as more moving objects are being detected. Static objects and background are labelled with \mbox{$l=0$}.

Ideally, for each detected object in frame $k$, the labels of all its points should be uniquely aligned with the labels of their correspondences in frame $k-1$.
However, in practice this is affected by the noise, image boundaries and occlusions.
To overcome this, we assign all the points with the label that appears most in their correspondences.
For a dynamic object, if the most frequent label in the previous frame is $0$, it means that the object starts to move, appears in the scene at the boundary, or reappears from occlusion. % a previously static object starts to move, or a new dynamic object enters the scene, or a previously dynamic and occluded object reappears.
In this case, the object is assigned a new tracking label.
\subsubsection{Object Motion Estimation}
\label{subsec:obj_mot_est}
As mentioned above, objects normally appear in small portions in the scene, which makes it hard to get sufficient sparse features to track and estimate their motions robustly.
We sample every third point within an object mask, and track them across frames.
Similar to the camera pose estimation, only inlier points are saved into the map and used for tracking in the next frame.
When the number of tracked object points decreases below a certain level, new object points are sampled and added.
We follow the same method as discussed in Section~\ref{sec:cam_pose} to generate an initial object motion model.
\subsection{Mapping}
In the mapping component, a global map is constructed and maintained.
Meanwhile, a local map is extracted from the global map, which is based on the current time step and a window of previous time steps.
Both maps are updated via a batch optimisation process.
\subsubsection{Local Batch Optimisation}
We maintain and update a local map. The goal of the local batch optimisation is to ensure accurate camera pose estimates are provided to the global batch optimisation. The camera pose estimation has a big influence on the accuracy of the object motion estimation and the overall performance of the algorithm.  
The local map is built using a fixed-size sliding window containing the information of the last $n_w$ frames, where $n_w$ is the window size and is set to $20$ in this paper.
Local maps share some common information; this defines the overlap between the different windows.
%We define a fixed-size overlap between the different windows that corresponds to the amount of common information shared by the local maps.
We choose to only locally optimise the camera poses and static structure within the window size, as locally optimising the dynamic structure does not bring any benefit to the optimisation unless a hard constraint (e.g. a constant object motion) is assumed within the window. However, the system is able to incorporate static and dynamic structure in the local mapping if needed.
When a local map is constructed, similarly, a factor graph optimisation is performed to refine all the variables within the local map, and then update them back into the global map.
\subsubsection{Global Batch Optimisation}
The output of the tracking component and the local batch optimisation consists of the camera pose, the object motions and the inlier structure.
These are saved in a global map that is constructed with all the previous time steps and is continually updated with every new frame.
A factor graph is constructed based on the global map after all input frames have been processed.
To effectively explore the temporal constraints, only points that have been tracked for more than $3$ instances are added into the factor graph.
The graph is formulated as an optimisation problem as described in Section~\ref{sec:graph_opt}.
The optimisation results serve as the output of the whole system.
\subsubsection{From Mapping to Tracking}
\label{sec:map_to_tracking}
Maintaining the map provides history information to the estimate of the current state in the tracking module, as shown in Fig.~\ref{fig:system_overview} with blue arrows going from the global map to multiple components in the tracking module of the system.
Inlier points from the last frame are leveraged to track correspondences in the current frame and estimate camera pose and object motions.
The last camera and object motion also serve as possible prior models to initialise the current estimation as described in Section~\ref{sec:cam_pose} and~\ref{subsec:obj_mot_est}.
Furthermore, object points help associate semantic masks across frames to ensure robust tracking of objects, by propagating their previously segmented masks in case of ``indirect occlusion'' resulting from the failure of semantic object segmentation.

%%%%%%%%%%%%%%%%%%%%%%%%%%%%%%%%%%%%%%%%%%%%%%%%%%%%%%%%%%%%%%%%%%%%%%%%%%%%%%%%%%%%%%%%%%%%%%%%%%%%%%%%%%%%%%%

\section{Experiments}
\label{sec:expe}
We evaluate VDO-SLAM in terms of camera motion, object motion and velocity, as well as object tracking performance.
The evaluation is done on the Oxford Multimotion Dataset~\cite{judd19ral} for indoor, and KITTI Tracking dataset~\cite{geiger13ijrr} for outdoor scenarios, with comparison to other state-of-the-art methods, including MVO~\cite{Judd18iros}, ClusterVO~\cite{huang2020cvpr}, DynaSLAM II~\cite{bescos2021ral} and CubeSLAM~\cite{Yang19tro}.
Due to the non-deterministic nature in running the proposed system, such as RANSAC processing, we run each sequence
$5$ times and take median values as the demonstrating results.
All the results are obtained by running the proposed system in default parameter setup.
Our open-source implementation includes the demo YAML files and instructions to run the system in both datasets.
% We also compare our results to three state-of-the-art methods, including MVO~\cite{Judd18iros}, ClusterVO~\cite{huang2020cvpr} and CubeSLAM~\cite{Yang19tro}, to prove the better performance of VDO-SLAM.

%
\subsection{Deep Model Setup}
We adopt a learning-based instance-level object segmentation, Mask R-CNN~\cite{He17iccv}, to generate object segmentation masks. The model of this method is trained on COCO dataset~\cite{lin14eccv}, and is directly used in this work without any fine-tuning.
For dense optical flow, we leverage a state-of-the-art method; PWC-Net~\cite{Sun18cvpr}.
The model is trained on FlyingChairs dataset~\cite{mayer16cvpr}, and then fine-tuned on Sintel~\cite{butler12eccv} and KITTI training datasets~\cite{Geiger12cvpr}.
To generate depth maps for a ``monocular'' version of our proposed system, we apply a learning-based monocular depth estimation method, MonoDepth2~\cite{godard19iccv}. The model is trained on Depth Eigen split~\cite{Eigen14nips} excluding the tested data in this paper.
Feature detection is done using FAST~\cite{rosten06eccv} implemented in~\cite{Rublee11iccv}.
All the above methods are applied using the default parameters.
\subsection{Error Metrics}
\label{sec:error-metrics}
We use a pose change error metric to evaluate the estimated $\SE$ motion, i.e., given a ground truth motion transform $\mathbf{T}$ and a corresponding estimated motion $\hat{\mathbf{T}}$, where \mbox{$\mathbf{T}\in \SE$} could be either a camera relative pose or an object motion.
The pose change error is computed as: \mbox{$\mathbf{E} = \hat{\mathbf{T}}^{-1}\:\mathbf{T}$}. 
This is similar to Relative Pose Error~\cite{sturm2012iros}, while we set the time interval $\Delta=1$ (per frame), because the trajectory of different object in a sequence varies from each other and are normally much shorter than the camera trajectory. \\
The translational error ${E}_{t}$ (meter) is computed as the $L_2$ norm of the translational component of $\mathbf{E}$.
The rotational error ${E}_{r}$ (degree) is calculated as the angle of rotation in an axis-angle representation of the rotational component of $\mathbf{E}$.
For different camera time steps and different objects in a sequence, we compute the root mean squared error (RMSE) for camera poses and object motions, respectively.
The object pose change in body-fixed frame is obtained by transforming the pose change $\prescript{0}{k-1}{\mathbf H}^{}_k$ in the inertial frame  into the body frame using the object pose ground-truth
\begin{align}
\prescript{L_{k-1}}{k-1}{\mathbf H}^{}_k
=
^{0}\mathbf{L}_{k-1}^{-1}  \
{\prescript{0}{k-1}{\mathbf H}^{}_k} \
{^{0}\mathbf{L}_{k-1}}.
\end{align}

We also evaluate the object speed error.
The linear velocity of a point on the object, expressed in the inertial frame, can be estimated by applying the pose change $\prescript{0}{k-1}{\mathbf{H}}_{k}$ and taking the difference
\begin{align}
\mathbf{v} \approx ^{0}{\mathbf{m}}_{k}^{i} - ^{0}{\mathbf{m}}_{k-1}^{i}
& = \left( \prescript{0}{k-1}{\mathbf{H}}^{}_{k} - \mathbf{I}_4 \right)
^{0}{\mathbf{m}}_{k-1}^{i} \notag \\
& = \prescript{0}{k-1}{\mathbf{t}}^{}_{k} - (\mathbf{I}_3 - \prescript{0}{k-1}{\mathbf{R}}^{}_{k})\: ^{0}\mathbf{m}^i_{k-1}.
\end{align}
To get a more reliable measurement, we average over all points on an object at a certain time.
Define $\mathbf{c}_{k-1} := \frac{1}{n} \sum \mathbf{m}_{k-1}^{i}$ for all n points on an object at time $k-1$.
Then
\begin{align}
\mathbf{v} & \approx \frac{1}{n} \sum_{i = 1}^{n}
\left( \prescript{0}{k-1}{\mathbf{t}}^{}_{k} - (\mathbf{I}_3 - \prescript{0}{k-1}{\mathbf{R}}^{}_{k})\: ^{0}\mathbf{m}^i_{k-1} \right)  \notag \\
& =  \prescript{0}{k-1}{\mathbf{t}}^{}_{k} - (\mathbf{I}_3 - \prescript{0}{k-1}{\mathbf{R}}^{}_{k})\: \mathbf{c}_{k-1}.
\end{align}
Then the speed error $E_s$ between the estimated $\hat{\mathbf{v}}$ and the ground truth $\mathbf{v}$ velocities can be calculated as: \mbox{$E_s = {|\hat{\mathbf{v}}|-|\mathbf{v}|}$}.
% For different objects tracked over temporal frames, we also compute the RMSE as an error metric.
%
\subsection{Oxford Multimotion Dataset}
\label{subsec:omd}
The recent Oxford Multimotion Dataset~\cite{judd19ral} contains sequences from a moving stereo or RGB-D camera sensor observing multiple swinging boxes or toy cars in an indoor scenario.
Ground truth trajectories of the camera and moving objects are obtained via a Vicon motion capture system.
We only choose the swinging boxes sequence ($500$ frames) for evaluation, since results of real driving scenarios are evaluated on KITTI dataset.
Note that, the trained model for instance segmentation cannot be applied to this dataset directly, since the training data (COCO) does not contain the class of square box. 
Instead, we use Otsu's method~\cite{otsu1979tsmc}, together with color information and multi-label processing to segment the boxes, which works very well for the simple setup of this dataset (color boxes that are highly distinguishable from the background). 
Table~\ref{tab:omd} shows results compared to the state-of-the-art MVO~\cite{Judd18iros} and ClusterVO~\cite{huang2020cvpr}, with data provided by the authors, respectively.
As they are both visual odometry systems without global refinement, we switch off the batch optimisation module in our system and generate our results for fair comparison.
We use the error metrics described in Section~\ref{sec:error-metrics}.
\begin{table*}[ht]
  \centering
  \fontsize{8}{9}\selectfont
  \caption{Comparison versus MVO~\cite{Judd18iros} and ClusterVO~\cite{huang2020cvpr} for camera pose and object motion estimation accuracy on the sequence of swinging\_4\_unconstrained sequence in Oxford Multi-motion dataset. Bold numbers indicate the better results.}
  \label{tab:omd}
 \begin{tabular}{ccccccc}
  \toprule
                           &\multicolumn{2}{c}{VDO-SLAM}  &\multicolumn{2}{c}{MVO}  &\multicolumn{2}{c}{ClusterVO}      \cr
  \midrule
                                            &$E_r$(deg) &$E_t$(m) &$E_r$(deg) &$E_t$(m)  &$E_r$(deg) &$E_t$(m) \cr
  \midrule
  Camera                                    &0.7709  &0.0112  &1.1948  &0.0314  &\textbf{0.7665}   &\textbf{0.0066}    \cr
  Top-left Swinging Box                &\textbf{1.1889}  &\textbf{0.0207}  &1.4553  &0.0288   &3.2537   &0.0673   \cr
  Top-right Swinging and rotating Box  &\textbf{0.7631}  &0.0132  &0.8992  &\textbf{0.0130}   &3.5308   &0.0256   \cr
  Bottom-left Swinging Box             &\textbf{0.9153}  &\textbf{0.0149}  &1.4949  &0.0261   &4.9146   &0.0763   \cr
  Bottom-right Rotating Box            &0.8469   &0.0192  &\textbf{0.7815}  &\textbf{0.0115}  &4.0675   &0.0144   \cr
  \bottomrule
 \end{tabular}
\end{table*}

Compared to MVO, our proposed method achieves better accuracy in the estimation of camera pose ($35\%$) and motion of the swinging boxes, top-left ($15\%$) and bottom-left ($40\%$).
We obtain slightly higher errors when there is spinning rotational motion of the object observed, in particular the top-right swinging and rotating box (in translation only), and the bottom-right rotating box.
We believe that this is due to using an optical flow algorithm that is not well optimised for self-rotating objects.
The consequence of this is poor estimation of point motion and consequent degradation of the overall object tracking performance.
Even with the associated performance loss for rotating objects, the benefits of dense optical flow motion estimation is clear in the other metrics.
Our method performs slightly worse than ClusterVO in the estimate of camera pose, and the translation of bottom-right rotating box.
Other than that, we achieve more than twice improvements against ClusterVO in the estimate of object motions.
% Overall, our proposed method achieves better accuracy than~\cite{Judd18iros} in $7$ out of $10$ error indexes for camera pose estimation and motion estimation of the $4$ moving boxes.
% In particular, our method achieves $50\%$ improvements in estimating the camera pose and motion of the swinging boxes, top-left and bottom-left.
% We obtain slightly higher errors when there is spinning rotational motion of the object observed, in particular the top-right swinging and rotating box, and the bottom-right rotating box.
% % Interestingly, our proposed algorithm performs worse than \cite{Judd18iros} in these cases as the algorithm is not designed for self-rotating motion and scenarios, but rather for the motion of relatively large objects outdoors, e.g. urban driving.
% We believe that this is due to using an optical flow algorithm that is not well optimised for self-rotating objects.
% The consequence of this is poor estimation of point motion and consequent degradation of the overall object tracking performance.
% Even with the associated performance loss for rotating objects, the benefits of dense optical flow motion estimation is clear in the other metrics. 
% %and improved CNN optical flow algorithms are under development in the computer vision literature.
%
\begin{figure}[ht]
 \centering
 \includegraphics[width=1.\columnwidth]{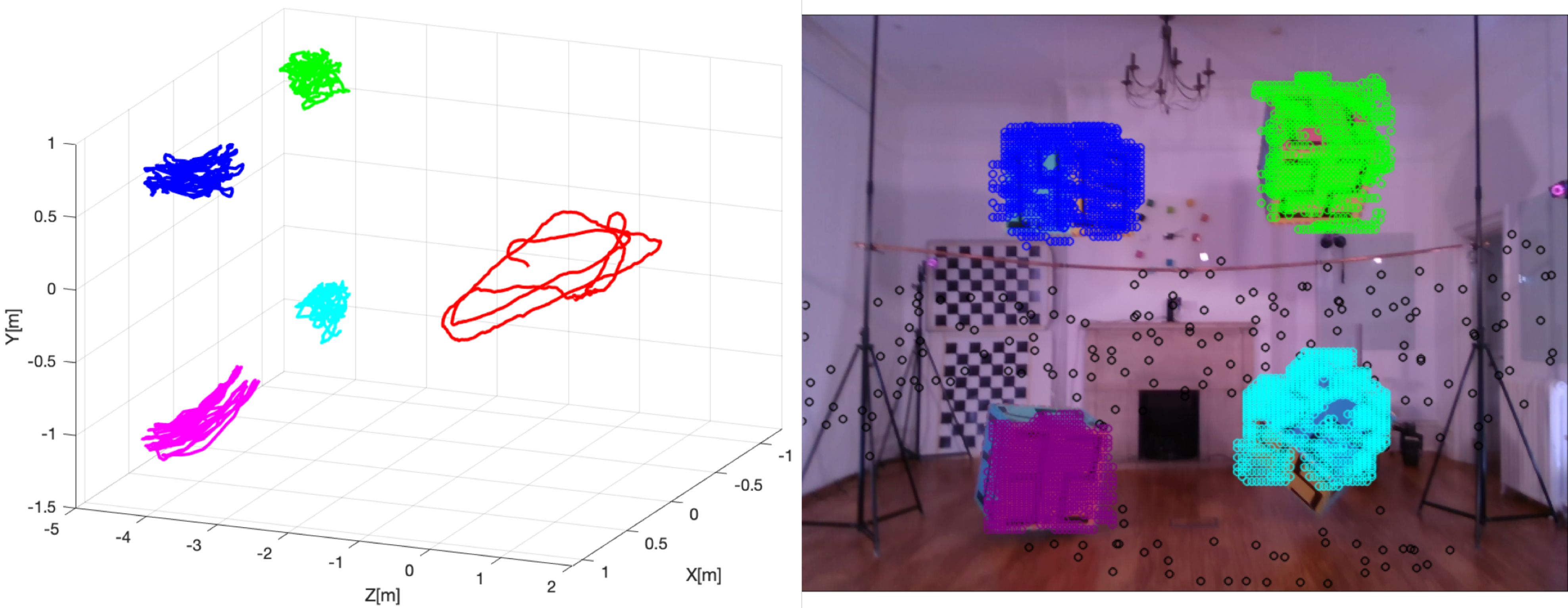}
 \caption{\textbf{Qualitative results of our method on Oxford Multimotion Dataset.} (Left) The $3$D trajectories of camera (red) and centres of the four boxes. (Right) Detected points on static background and object body. Black color corresponds to static points and features on each object are shown in a different color.}
\label{fig:oxford_demo}
\end{figure}

An illustrative result of the trajectory output of our algorithm on Oxford Multimotion Dataset is shown in Fig.~\ref{fig:oxford_demo}.
Tracks of dynamic features on swinging boxes visually correspond to the actual motion of the boxes.
This can be clearly seen in the swinging motion of the bottom-left box shown with purple color in Fig.~\ref{fig:oxford_demo}.
%The top right box is swinging and rotating which makes its motion visualisation using points only a bit harder.
%
\subsection{KITTI Tracking Dataset}
\label{subsec:kitti}
The KITTI Tracking Dataset~\cite{geiger13ijrr} contains $21$ sequences in total with ground truth information about camera and object poses. Among these sequences, some are not included in the evaluation of our system; as they contain no moving objects (static only scenes) or only contain pedestrians that are non-rigid objects, which is outside the scope of this work.
Note that, as only rotation around Y-axis is provided in the ground truth object poses, we assign zeros to the other two axes for the convenience of full motion evaluation.
\begin{table*}[t]
  \centering
  \caption{Comparison versus DynaSLAM II~\cite{bescos2021ral} and CubeSLAM~\cite{Yang19tro} for camera pose and object motion estimation accuracy on nine sequences with moving objects drawn from the KITTI dataset.
  Bold numbers indicate the better result.} %underlined bold numbers indicate an order of magnitude improvement.}
  \label{tab:kitti}
  \fontsize{8}{9}\selectfont
  \begin{tabular}{ccccccccccccccc}
  \toprule
     &\multicolumn{2}{c}{DynaSLAM II} &\multicolumn{4}{c}{VDO-SLAM (RGB-D)} & \multicolumn{4}{c}{VDO-SLAM (Monocular)}  &\multicolumn{4}{c}{CubeSLAM}       \cr
  \midrule
     &\multicolumn{2}{c}{Camera}
     &\multicolumn{2}{c}{Camera} &\multicolumn{2}{c}{Object}
     &\multicolumn{2}{c}{Camera} &\multicolumn{2}{c}{Object}
     &\multicolumn{2}{c}{Camera} &\multicolumn{2}{c}{Object} \cr
  Seq &$E_r$(deg) &$E_t$(m)  &$E_r$(deg) &$E_t$(m) &$E_r$(deg) &$E_t$(m) &$E_r$(deg) &$E_t$(m) &$E_r$(deg) &$E_t$(m) &$E_r$(deg) &$E_t$(m) &$E_r$(deg) &$E_t$(m) \cr
  \midrule
  00   &\textbf{0.06} &\textbf{0.04} &0.0741  &0.0674  &\textbf{1.0520}  &\textbf{0.1077} &0.1830 &0.1847 &2.0021 &0.3827 &-  &-  &-  &- \cr
  01   &0.04 &\textbf{0.05} &\textbf{0.0382}  &0.1220  &\textbf{0.9051}  &\textbf{0.1573} &0.1772 &0.4982 &1.1833 &0.3589 &-  &-  &-  &-  \cr
  02   &0.02 &\textbf{0.04} &\textbf{0.0182}  &0.0445  &\textbf{1.2359}  &\textbf{0.2801} &0.0496 &0.0963 &1.6833 &0.4121 &-  &-  &- &-  \cr
  03   &0.04 &\textbf{0.06} &\textbf{0.0311}  &0.0816  &\textbf{0.2919}  &\textbf{0.0965} &0.1065 &0.1505 &0.4570 &0.2032 & 0.0498 & 0.0929 & 3.6085 & 4.5947 \cr
  04   &0.06 &\textbf{0.07} &\textbf{0.0482}  &0.1114  &\textbf{0.8288}  &\textbf{0.1937} &0.1741 &0.4951 &3.1156 &0.5310 & 0.0708 & 0.1159 & 5.5803 & 32.5379 \cr
  05   &0.03 &\textbf{0.06} &\textbf{0.0219}  &0.0932  &\textbf{0.3705}  &\textbf{0.1140} &0.0506 &0.1368 &0.6464 &0.2669 & 0.0342 &0.0696 & 3.2610 & 6.4851 \cr
  06   &\textbf{0.04} &0.02 &0.0488  &\textbf{0.0186}  &\textbf{1.0803}  &\textbf{0.1158} &0.0671 &0.0451 &2.0977 &0.2394 &-  &-  &-  &-  \cr
  18   &\textbf{0.02} &\textbf{0.05} &0.0211  &0.0749  &\textbf{0.2453}  &\textbf{0.0825} &0.1236 &0.3551 &0.5559 &0.2774  & 0.0433 &0.0510 & 3.1876 & 3.7948 \cr
  20   &0.04 &\textbf{0.07} &\textbf{0.0271}  &0.1662  &\textbf{0.3663}  &\textbf{0.0824} &0.3029 &1.3821 &1.1081 &0.3693  & 0.1348 & 0.1888 & 3.4206 & 5.6986 \cr
  \bottomrule
  \end{tabular}
\end{table*}
\begin{figure}[ht]
 \centering
 \includegraphics[width=1.0\columnwidth]{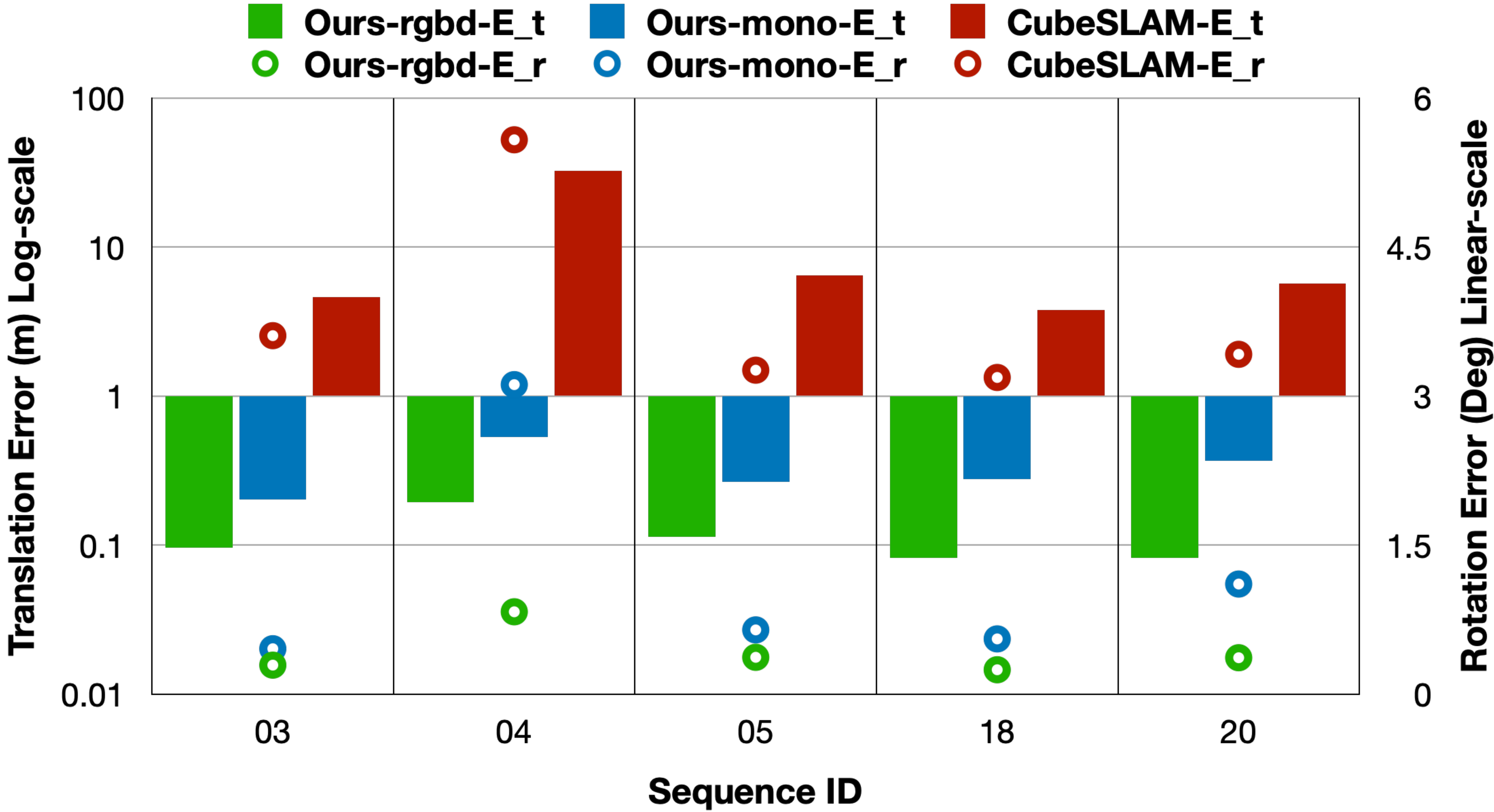}
 \caption{\textbf{Accuracy of object motion estimation of our method compared to CubeSLAM (\cite{Yang19tro}).} The color bars refer to translation error that is corresponding to the left Y-axis in log-scale. The circles refer to rotation error, which corresponds to the right Y-axis in linear-scale.}
\label{fig:obj_mot_err}
\end{figure}
\subsubsection{Camera Pose and Object Motion}
Table~\ref{tab:kitti} demonstrates results of both camera pose and object motion estimation in nine sequences, compared to DynaSLAM II~\cite{bescos2021ral} and CubeSLAM~\cite{Yang19tro}. 
Results of DynaSLAM II is obtained directly from their paper, where only the evaluation of camera pose is available.
We initially tried to evaluate CubeSLAM ourselves with the default provided parameters, however errors were much higher, and hence we only report results of five sequences provided by the authors of CubeSLAM  after some correspondences.
As CubeSLAM is for monocular camera, we also compute results of a learning-based monocular version of our proposed method (as mentioned in Section~\ref{sec:system}) for fair comparison.

% Overall, both our proposed RGB-D and learning-based monocular methods obtain high accuracy in both camera and object motion estimation.
Our proposed method achieves competitive and high accuracy in comparison with DynaSLAM II for the estimate of camera pose. 
In particular, our method obtains slightly lower rotational errors while higher translational errors than DynaSLAM II. 
We believe the difference in accuracy is due to the underlying formulations in estimating camera pose. 
% We believe this is due to the different formulations in estimating camera pose, i.e., our global optimization is performed in $3$D space, which better constrains the distant points that are good for rotation estimation, while DynaSLAM II formulates the optimization on image plane that is less sensitive to depth noise and good for translation estimation. 
When compared to CubeSLAM, our RGB-D version gets lower errors in camera pose, while our learning-based monocular version slightly higher.
We believe the weak performance of monocular version is because the model does not capture the scale of depth accurately with only monocular input. 
Nevertheless, both versions obtain consistently lower errors in object motion estimation.
In particular, as demonstrated in Fig.~\ref{fig:obj_mot_err}, the translation and rotation errors in CubeSLAM are all above $3$ meters and $3$ degrees, with errors reaching $32$ meters and $5$ degrees in extreme cases respectively.
However, our translation errors vary between $0.1$-$0.3$ meters and rotation errors between $0.2$-$1.5$ degrees in the case of RGB-D, and $0.1$-$0.3$ meters, and $0.4$-$3.1$ degrees in the case of learning-based monocular, which indicates that our object motion estimation achieves an order of magnitude improvements in most cases.
In general, the results suggest that point-based object motion/pose estimation methods is more robust and accurate than those using high-level geometric models, probably due to the fact that geometric model extraction could lead to losing information and introducing more uncertainty. 
%It is also worth mentioning that when we tried to evaluate CubeSLAM ourselves with the default provided parameters, errors were a lot worse, which proves the sensitivity of the algorithm to the chosen set of parameters, unlike our method where only one set of parameters was chosen for all the reported experiments in this paper.
%
\subsubsection{Object Tracking and Velocity}
\begin{figure}[ht]
 \centering
 \includegraphics[width=1.0\columnwidth]{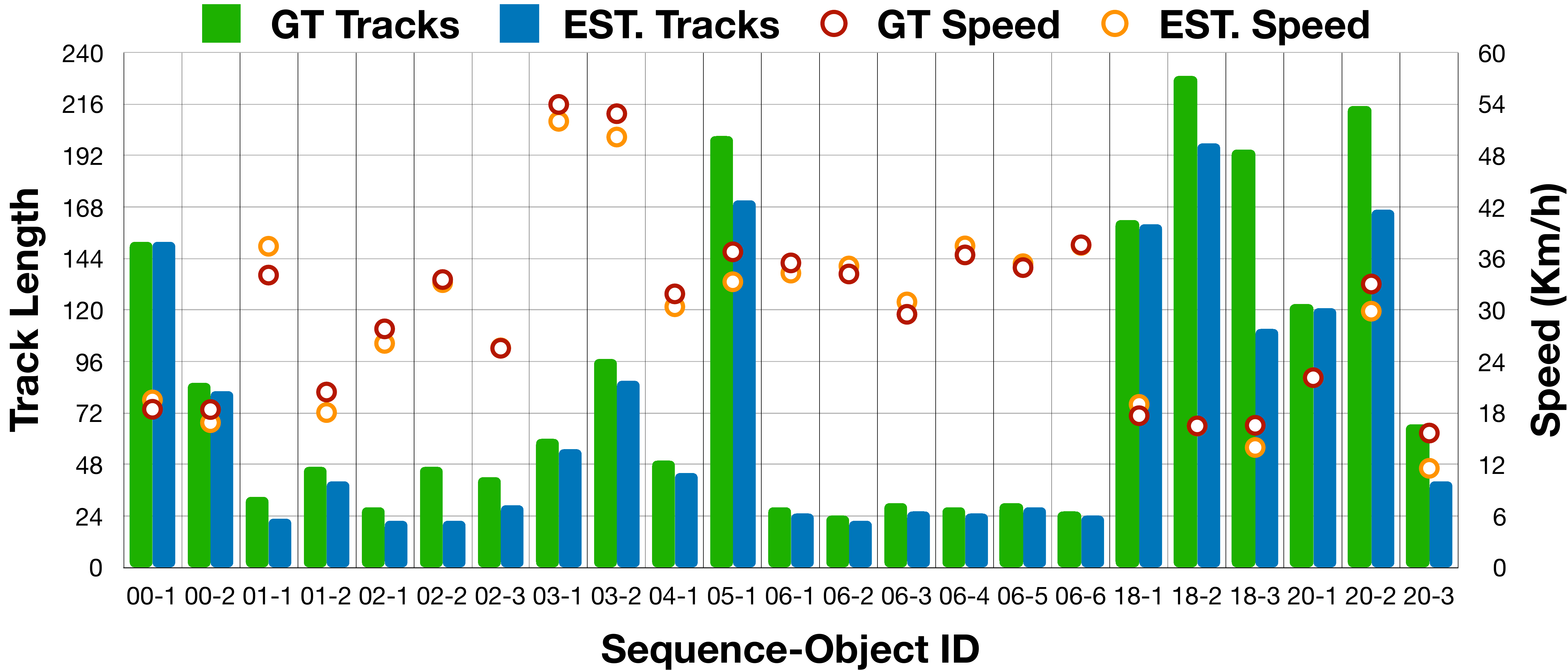}
 \caption{\textbf{Tracking performance and speed estimation.} Results of object tracking length and object speed for some selected objects (tracked for over $20$ frames), due to limited space. The color bars represent the length of object tracks, which is corresponding to the left Y-axis. The circles represent object speeds, which is corresponding to the right Y-axis. GT refers to ground truth, and EST. refers to estimated values.}
\label{fig:obj_track}
\end{figure}

We also demonstrate the performance of tracking dynamic objects, and show results of object speed estimation, which is an important information for autonomous driving applications.
Fig.~\ref{fig:obj_track} illustrates results of object tracking length and object speed for some selected objects (tracked for over $20$ frames) in all the tested sequences.
Our system is able to track most objects for more than $80\%$ of their occurrence in the sequence. Moreover, our estimated objects speed is always consistently close to the ground truth.
\subsubsection{Qualitative Results}
\begin{figure*}[ht]
	\begin{minipage}[t]{0.45\textwidth}
		\begin{minipage}[b]{\textwidth}
			\centering
			\includegraphics[width=1\linewidth]{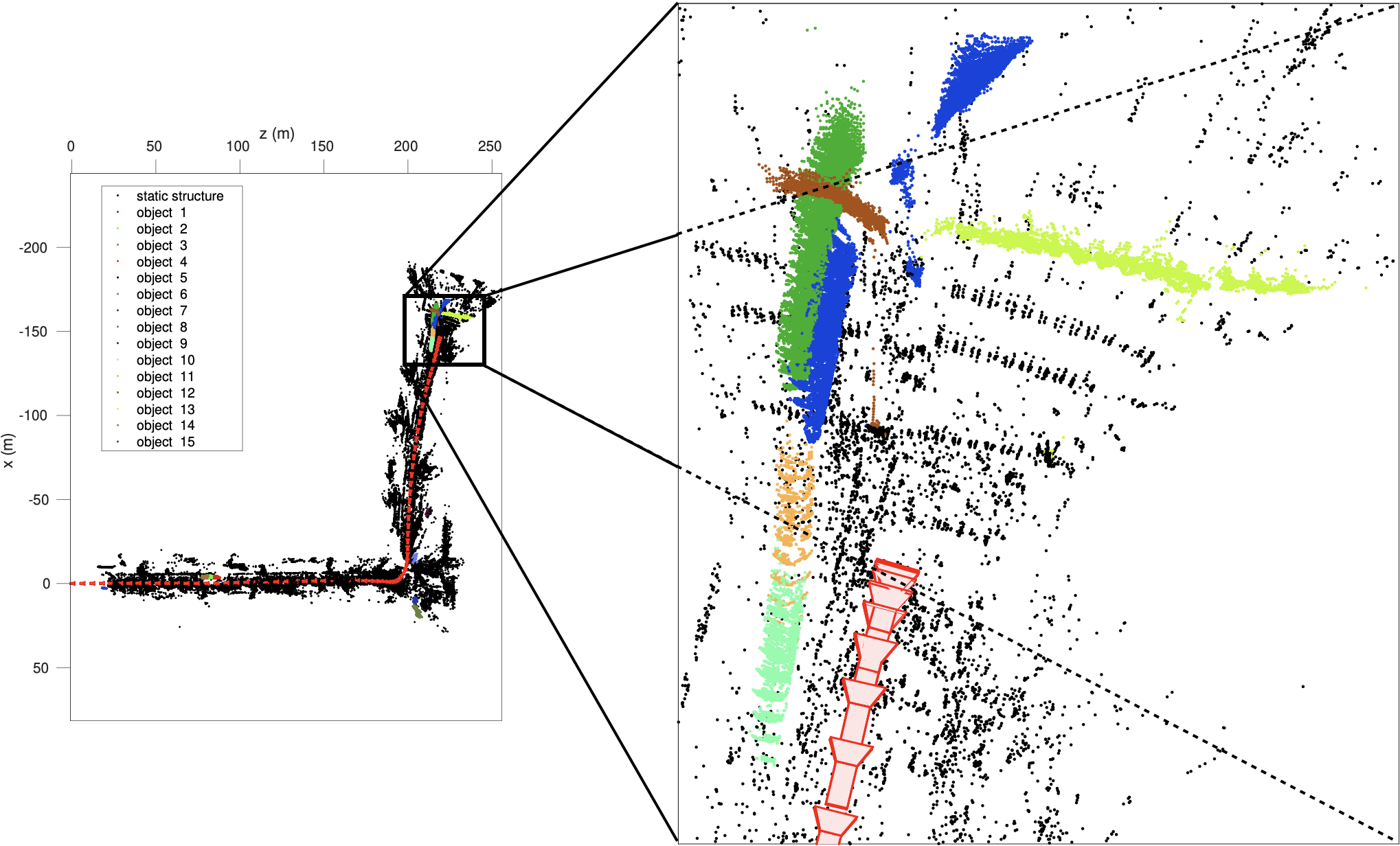}
			% (a) \scriptsize Seq. 01 and a zoom-in at the intersection at the end of the sequence.
		\end{minipage}
		\label{fig:kitti1}
	\end{minipage}\hspace{.8cm}
	\begin{minipage}[t]{0.34\textwidth}
		\begin{minipage}[b]{\textwidth}
			\centering
			\includegraphics[width=1\linewidth]{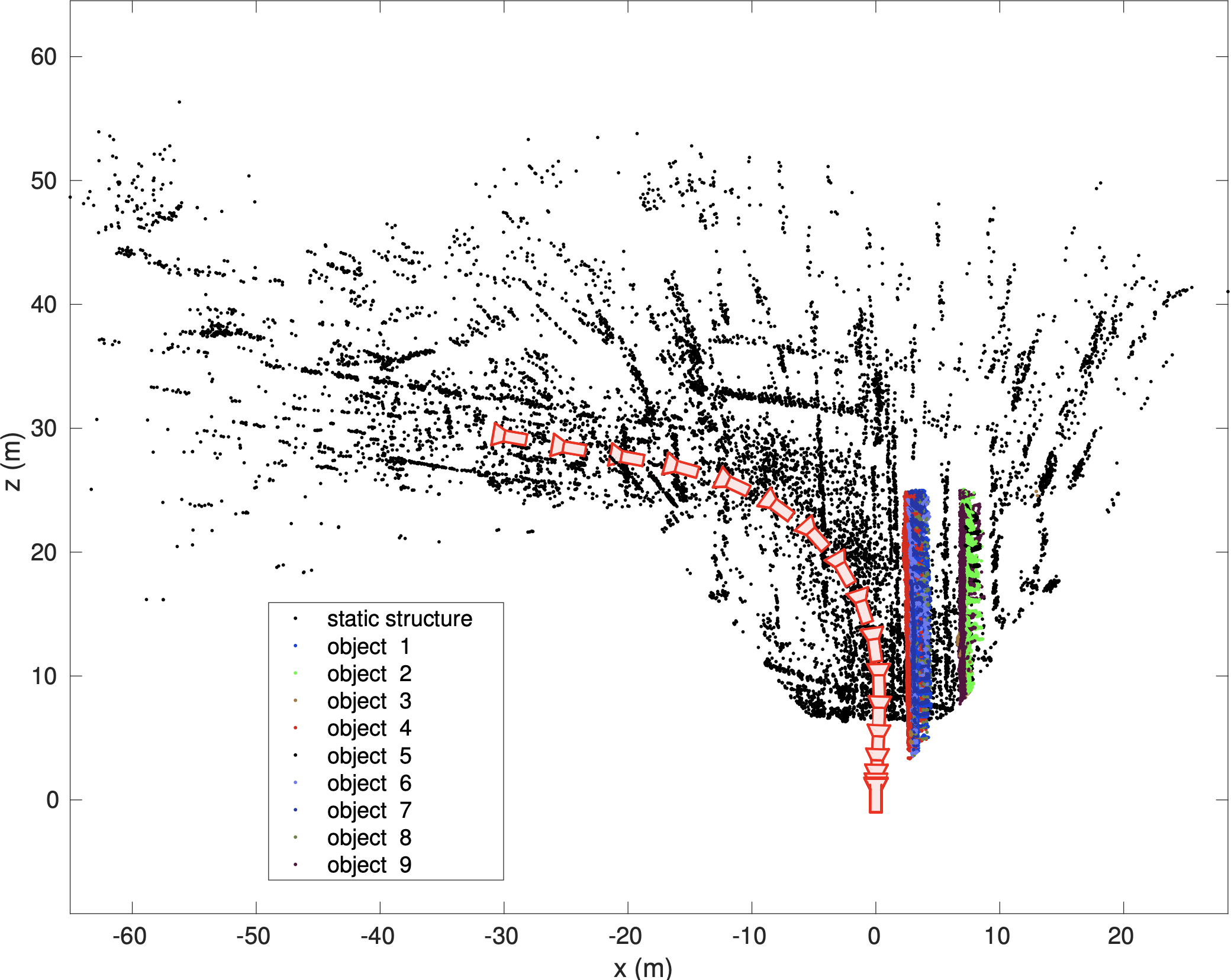}
			% \\(b) \scriptsize Seq. 06
		\end{minipage}
		\label{fig:kitti6}
	\end{minipage}
	\centering
	\begin{minipage}[t]{0.8\textwidth}
		\begin{minipage}[b]{\textwidth}
			\centering
			\includegraphics[width=1\linewidth]{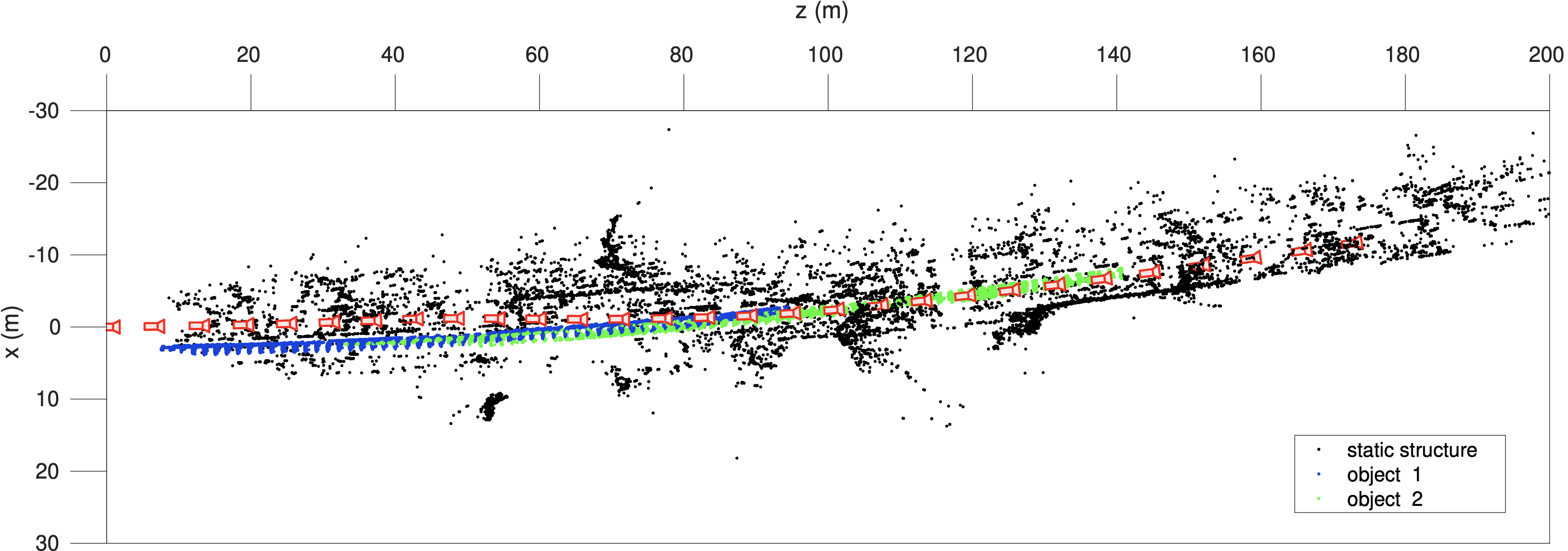}
			% \\(c) \scriptsize Seq. 03
		\end{minipage}
		\label{fig:kitti3}
	\end{minipage}
	\caption{\textbf{Illustration of system output; a dynamic map with camera poses, static background structure, and tracks of dynamic objects.} Sample results of VDO-SLAM on KITTI sequences. Black represents static background, and each detected object is shown in a different colour. Top left figure represents Seq.$01$ and a zoom-in on the intersection at the end of the sequence, top right figure represents Seq.$06$ and bottom figure represents Seq.$03$.}
	\label{fig:kitti}
\end{figure*}
Fig.~\ref{fig:kitti} illustrates the output of our system for three of the KITTI sequences.
The proposed system is able to output the camera poses, along with the static structure and dynamic tracks of every detected moving object in the scene in a spatiotemporal map representation.

\subsection{Discussion}
\label{subsec:discuss}
Apart from the extensive evaluation in Section~\ref{subsec:kitti} and~\ref{subsec:omd}, we also provide detailed experimental results to prove the effectiveness of key modules in our proposed system.
Finally, the computational cost of the proposed system is discussed.
\begin{table}[ht]
  \centering
  \caption{The number of points tracked for more than five frames on the nine sequences of the KITTI dataset. Bold numbers indicate the better results. Underlined bold numbers indicate an order of magnitude increase in number.}
  \label{tab:track_point}
  \fontsize{8}{9}\selectfont
  \begin{tabular}{ccccc}
  \toprule
     &\multicolumn{2}{c}{Background} &\multicolumn{2}{c}{Object} \cr
  Seq   &Motion Only &Joint &Motion Only &Joint  \cr
  \midrule
  00    &1798 &\textbf{\underline{12812}} &1704 &\textbf{7162}  \cr
  01    &237 &\textbf{\underline{5075}}   &907 &\textbf{4583}     \cr
  02    &7642 &\textbf{10683} &52 &\textbf{\underline{1442}}    \cr
  03    &778 &\textbf{\underline{12317}}  &343 &\textbf{\underline{3354}}    \cr
  04    &9913 &\textbf{25861} &339 &\textbf{\underline{2802}}   \cr
  05    &713 &\textbf{\underline{11627}}  &2363 &\textbf{2977}   \cr
  06    &7898 &\textbf{11048} &482 &\textbf{\underline{5934}}   \cr
  18    &4271 &\textbf{22503} &5614 &\textbf{14989} \cr
  20    &9838 &\textbf{49261} &9282 &\textbf{13434} \cr
  \bottomrule
  \end{tabular}
\end{table}

\subsubsection{Robust Tracking of Points}
The graph optimisation explores the spacial and temporal information to refine the camera poses and the object motions, as well as the static and dynamic structure.
This process requires robust tracking of good points in terms of both quantity and quality.
This was achieved by refining the estimated optical flow jointly with the motion estimation, as discussed in Section~\ref{sec:FlowRefine}.
The effectiveness of joint optimisation is shown by comparing a baseline method that only optimises for the motion (Motion Only) using~\eqref{eq:camera_pose_2D_cost} for camera motion or~\eqref{eq:obj_mot_2D_cost} for object motion, and the improved method that optimises for both the motion and the optical flow (Joint) using~\eqref{eq:cam_pose_flow_2D_cost} or~\eqref{eq:obj_mot_flow_2D_cost}.
Table~\ref{tab:track_point} demonstrates that the joint method obtains considerably more points that are tracked for long periods.
\begin{table}[ht]
  \centering
  \caption{Average camera pose and object motion errors over the nine sequences of the KITTI dataset. Bold numbers indicate the better results.}
  \label{tab:avg_err}
  \fontsize{8}{9}\selectfont
  \begin{tabular}{ccccc}
  \toprule
     &\multicolumn{2}{c}{Motion Only} &\multicolumn{2}{c}{Joint}  \cr
     &$E_r$(deg) &$E_t$(m) &$E_r$(deg) &$E_t$(m)  \cr
  \midrule
  Camera    &0.0412 &0.0987 &\textbf{0.0365} &\textbf{0.0866}     \cr
  Object    &1.0179 &0.1853 &\textbf{0.7085} &\textbf{0.1367}     \cr
  \bottomrule
  \end{tabular}
\end{table}

Using the tracked points given by the joint estimation process leads to better estimation of both camera pose and object motion.
As demonstrated in Table~\ref{tab:avg_err}, an improvement of about $10\%$ (camera) and $25\%$ (object) in both translation and rotation errors was observed over the nine sequences of the KITTI dataset shown above.

\subsubsection{Robustness against Non-direct Occlusion}
\begin{figure}[ht]
	\begin{minipage}[h]{0.45\textwidth}
		\begin{minipage}[b]{\textwidth}
			\centering
			\includegraphics[width=1\linewidth]{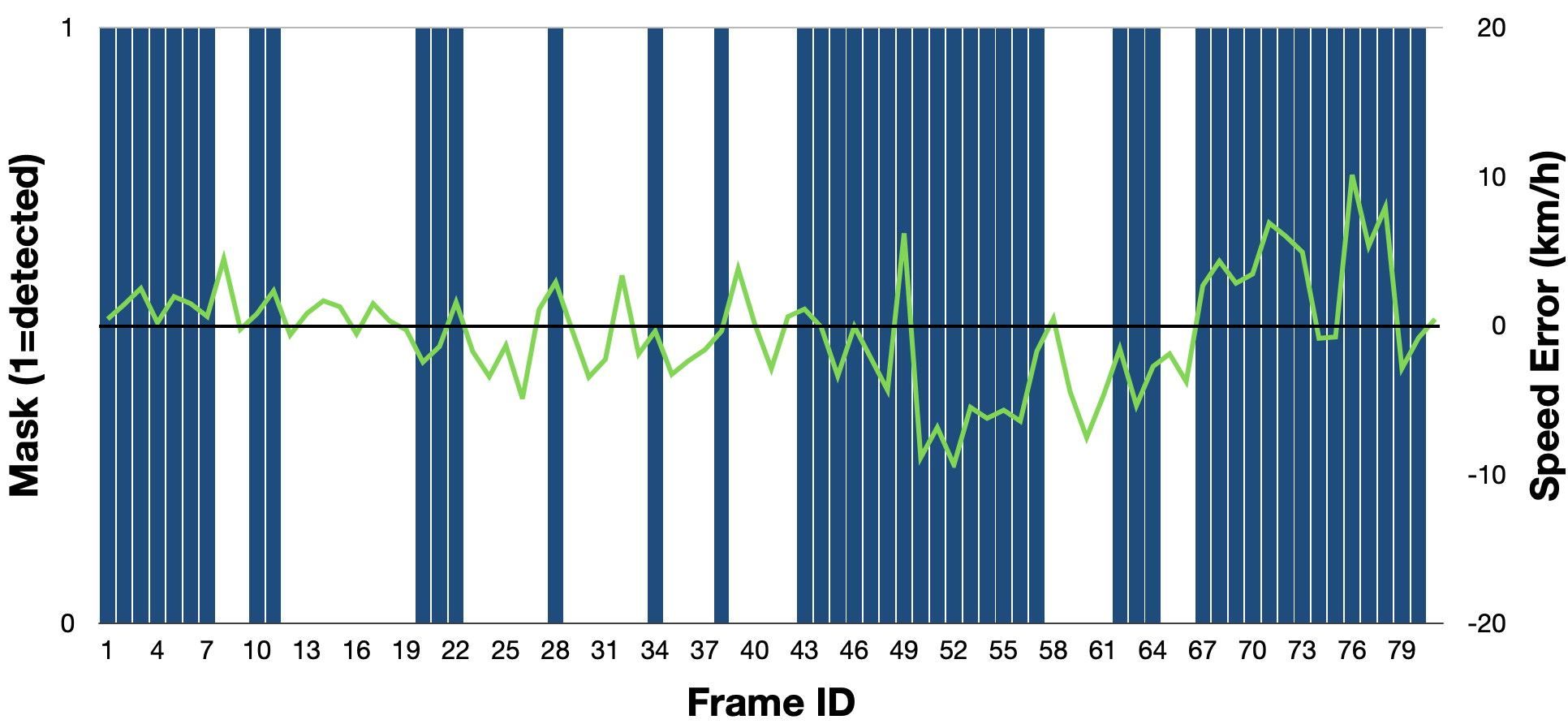}
		\end{minipage}
		\label{fig:fail_mask}
	\end{minipage}
	\begin{minipage}[h]{0.22\textwidth}
		\begin{minipage}[b]{\textwidth}
			\centering
			\includegraphics[width=1\linewidth]{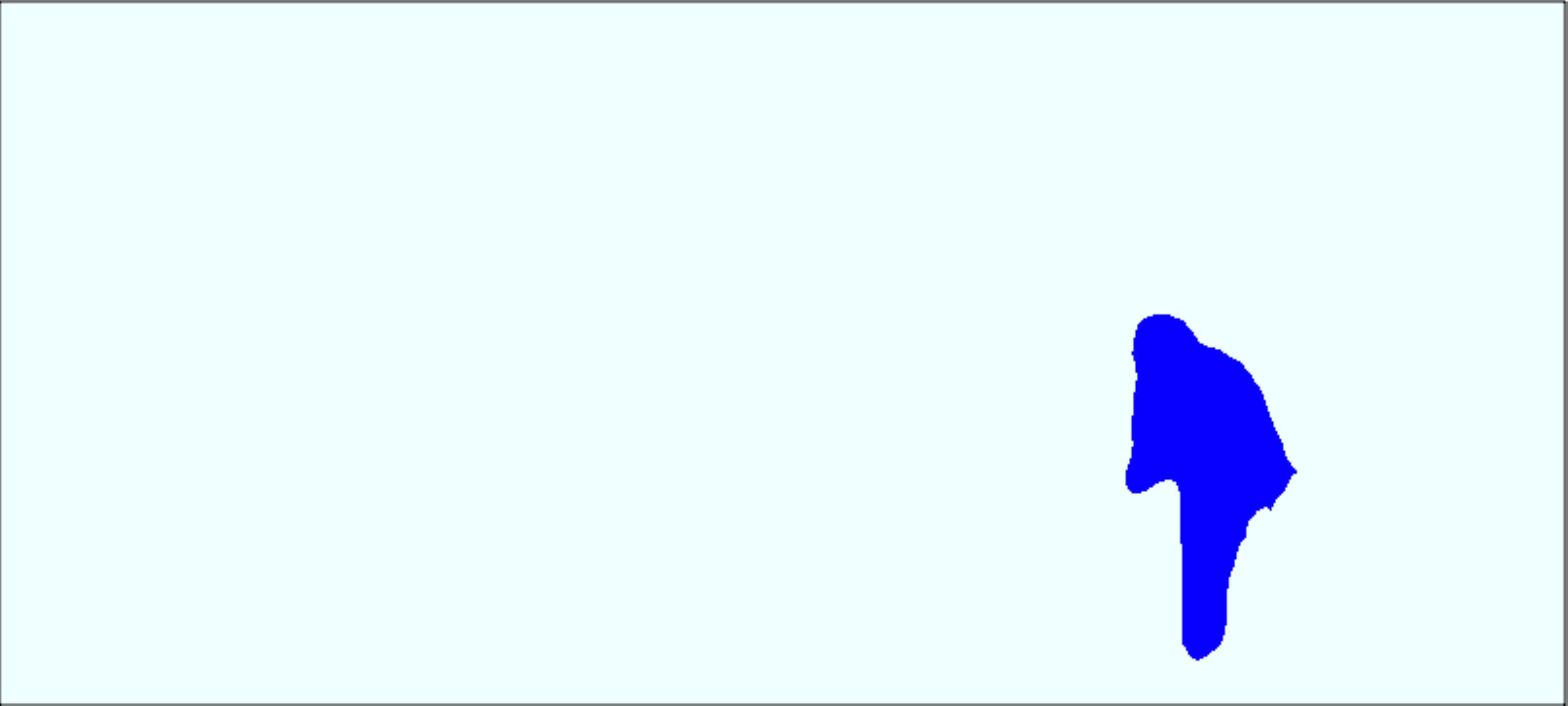}
		\end{minipage}
		\label{fig:obj_mask}
	\end{minipage}
	\centering
	\begin{minipage}[h]{0.22\textwidth}
		\begin{minipage}[b]{\textwidth}
			\centering
			\includegraphics[width=1\linewidth]{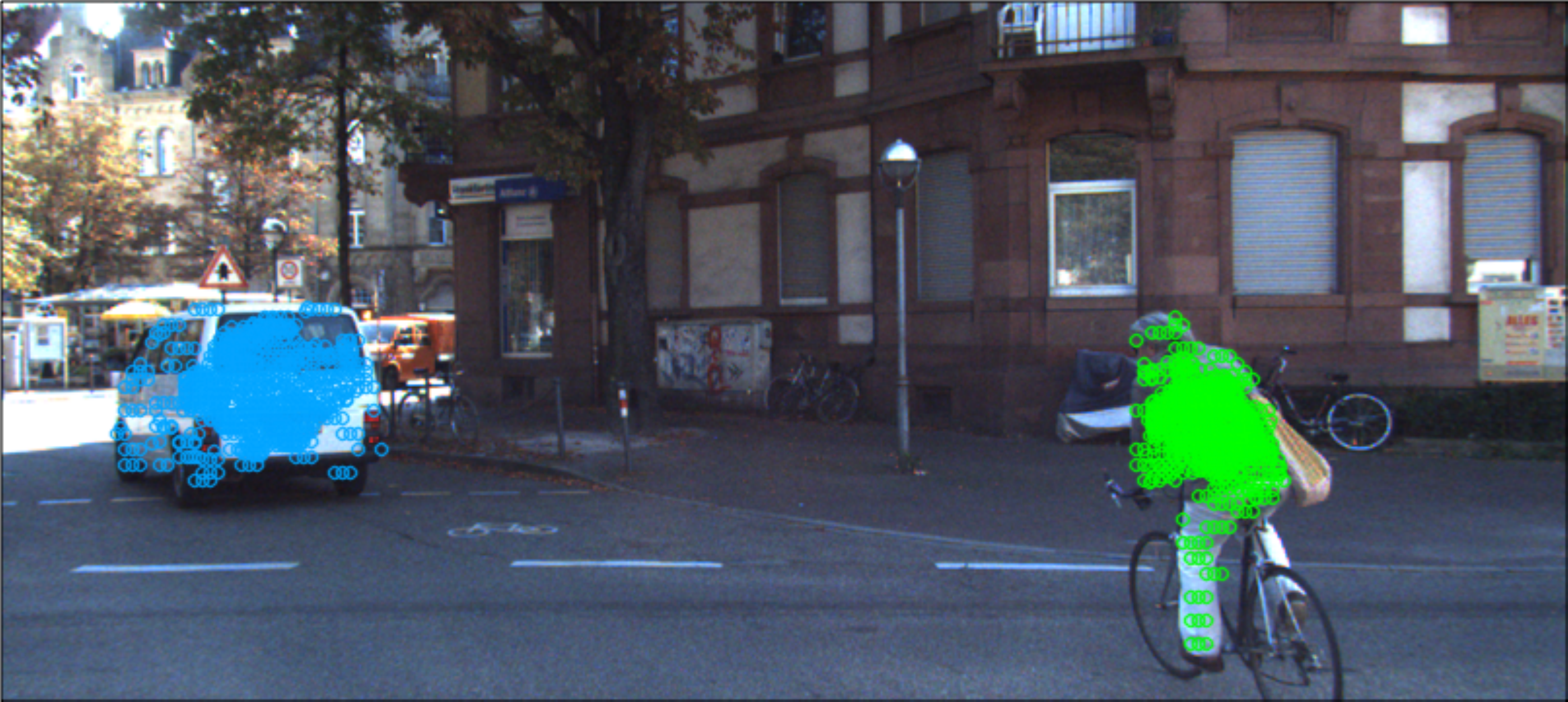}
		\end{minipage}
		\label{fig:fea_track}
	\end{minipage}
	\caption{\textbf{Robustness in tracking performance and speed estimation in case of semantic segmentation failure.}\\ An example of tracking performance and speed estimation for a white van (ground-truth average speed $20$km/h) in Seq.$00$. (Top) Blue bars represent a successful object segmentation, and green curves refer to the object speed error. (Bottom-left) An illustration of semantic segmentation failure on the van. (Bottom-right) Result of propagating the previously tracked features on the van by our system.}
	\label{fig:tracking_performance}
\end{figure}

The mask segmentation may fail in some cases, due to direct or indirect occlusions (illumination change, etc.).
Thanks to the mask propagating method described in Section~\ref{sec:map_to_tracking}, our proposed system is able to handle mask failure cases caused by indirect occlusions.
Fig.~\ref{fig:tracking_performance} demonstrates an example of tracking a white van for $80$ frames, where the mask segmentation fails in $33$ frames.
Despite the object segmentation failure, our system is still continuously able to track the van, and estimate its speed with an average error of $2.64$ km/h across the whole sequence. Speed errors in the second half of the sequence are higher due to partial direct occlusions, and increased distance to the object getting farther away from the camera.
\subsubsection{Global Refinement on Object Motion}
\begin{figure}[ht]
	\begin{minipage}[t]{0.48\textwidth}
	\centering
		\begin{minipage}[b]{\textwidth}
			% \centering
			\includegraphics[width=1\linewidth]{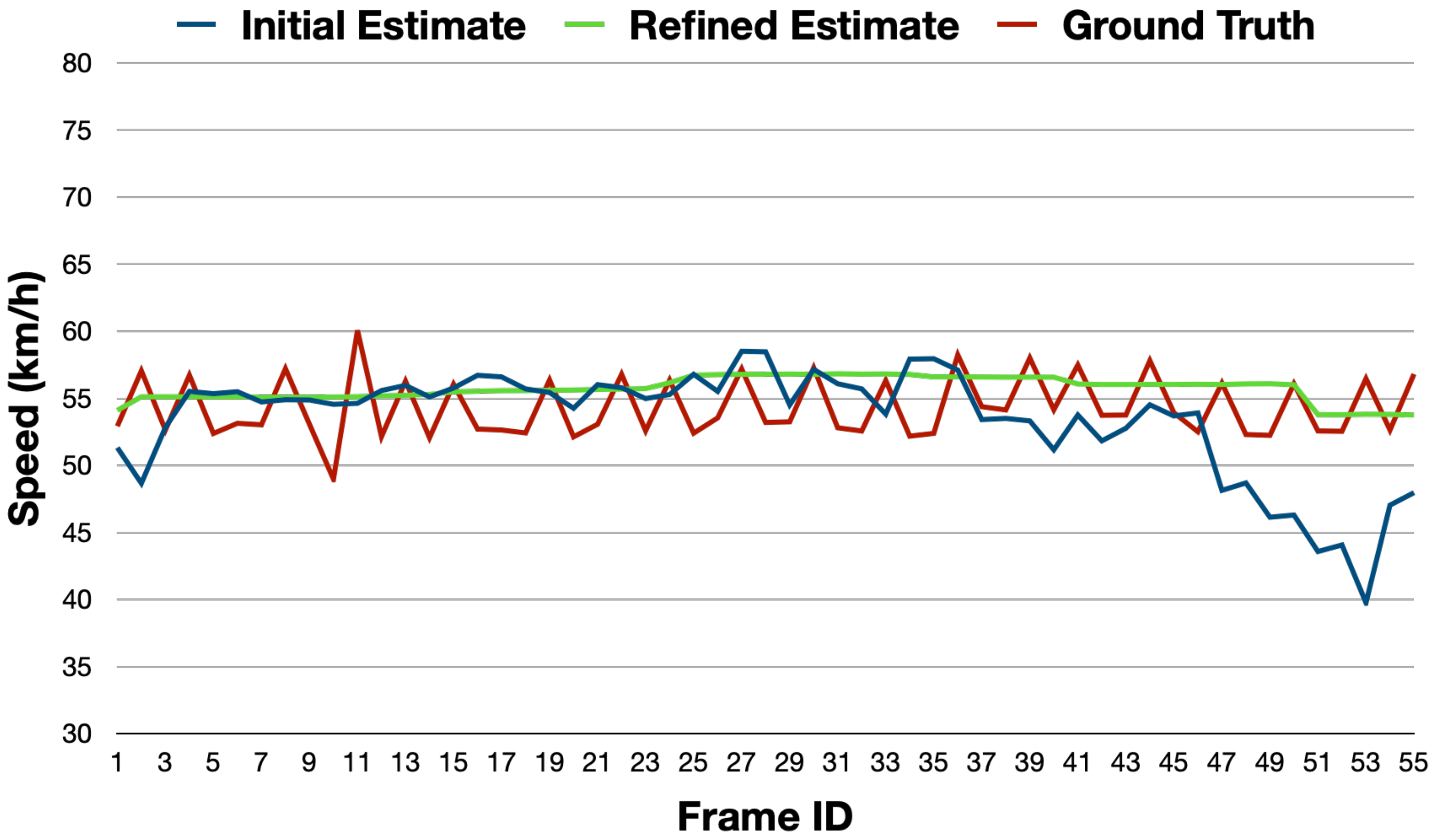}
		\end{minipage}
	\end{minipage}
	% \centering
	% \begin{minipage}[t]{0.45\textwidth}
	% 	\begin{minipage}[b]{\textwidth}
	% 		% \centering
	% 		\includegraphics[width=1\linewidth]{improve_ratio.pdf}
	% 	\end{minipage}
	% \end{minipage}
	\caption{\textbf{Global refinement effect on object speed estimation.} The initial (blue) and refined (green) estimated speeds of a wagon in Seq.$03$, travelling along a straight road, compared to the ground truth speed (red). Note the ground truth speed is slightly fluctuating. We believe it is due to the ground truth object poses being approximated from lidar scans.}
	\label{fig:glo_refine}
\end{figure}

Initial object motion estimation (in the tracking component of the system) is independent between frames, since it is purely related to the sensor measurements.
As illustrated in Fig.~\ref{fig:glo_refine}, the blue curve describes an initial object speed estimate of a wagon observed for $55$ frames in sequence $03$ of the KITTI tracking dataset.
As seen in the figure, the speed estimation is not smooth and large errors occur towards the second half of the sequence.
This is mainly caused by the increased distance to the object getting farther away from the camera, and its structure only occupying a small portion of the scene.
In this case, the object motion estimation from sensor measurements solely becomes challenging and error-prone.
Therefore, we formulate a factor graph and refine the motions together with the static and dynamic structure as discussed in Section~\ref{sec:graph_opt}.
The green curve in Fig.~\ref{fig:glo_refine} shows the object speed results after the global refinement, which becomes smoother in the first half of the sequence and is significantly improved in the second half.

\begin{figure}[ht]
 \centering
 \includegraphics[width=1.0\columnwidth]{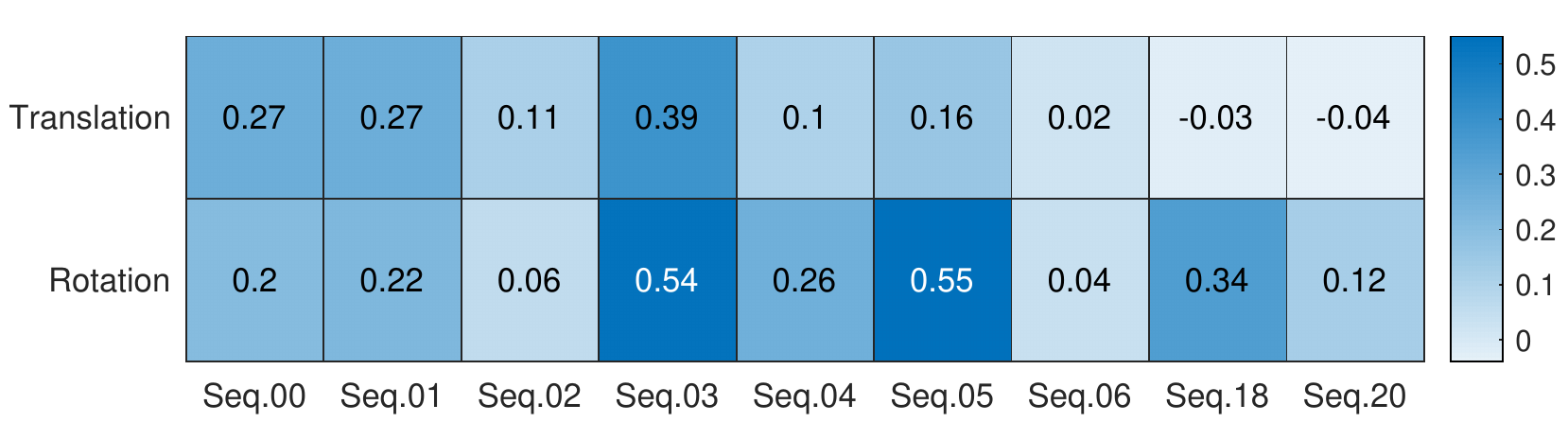}
 \caption{\textbf{Improvement on object motion after graph optimization.} The numbers in the heatmap show the ratio of decrease in error on the nine sequences of the KITTI dataset.}
\label{fig:improve_ratio}
\end{figure}

Fig.~\ref{fig:improve_ratio} demonstrates the average improvement for all objects in each sequence of KITTI dataset. 
With graph optimization, the errors can be reduced up to $39\%$ in translation and $55\%$ in rotation.
Interestingly, the translation errors in Seq.$18$ and Seq.$20$ increase slightly.
We believe it is because the vehicles keep alternating between acceleration and deceleration due to the heavy traffic jams in both sequences, which strongly violates the smooth motion constraint that is set for general cases.
\subsubsection{Computational Analysis}
Finally, we provide the computational analysis of our system. The experiments are carried out on an Intel Core i$7$ $2.6$ GHz laptop computer with $16$ GB RAM.
The object semantic segmentation and dense optical flow computation times depend on the GPU power and the CNN model complexity.
Many current state-of-the-art algorithms can run in real time (\cite{Bolya19iccv, Hui20tpami}). In this paper, the semantic segmentation and optical flow results are produced off-line as input to the system.
The SLAM system is implemented in C++ on CPU using a modified version of g$2$o as a back-end~\cite{Kummerle11icra}. We show the computational time in Table~\ref{tab:runtime} for both datasets.
% In the local batch optimisation, the window size is set to $20$ frames with an overlap of $4$ frames. The time cost of every system component is averaged over all frames, and sequences.
Overall, the tracking part of our proposed system is able to run at the frame rate of \mbox{$5$-$8$} fps depending on the number of detected moving objects, which can be improved by employing parallel implementation.
The runtime of the global batch optimisation strongly depends on the amount of camera poses (number of frames), and objects (density in terms of the number of dynamic objects observed per frame) present in the scene.
\begin{table}[ht]
  % \vspace{-.2cm}
  \centering
  \fontsize{8}{9}\selectfont
  \caption{Runtime of different system components for both datasets. The time cost of every component is averaged over all frames and sequences, except for the object motion estimation and object motion estimation that are averaged over the number of objects.}
  \label{tab:runtime}
 \begin{tabular}{ccc}
  \toprule
  Dataset                &Tasks                       &Runtime (mSec)                      \cr
  \midrule
  \multirow{5}{*}{KITTI}
                         &Feature Detection           &16.2550                             \cr
                         &Camera Pose Estimation      &52.6542                             \cr
                         &Dynamic Object Tracking (avg/object)    &8.2980                             \cr
                         &Object Motion Estimation (avg/object)    &22.9081                             \cr
                         &Map and Mask Updating       &22.1830                             \cr
                         &Local Batch Optimisation    &18.2828                \cr
  \midrule
  \multirow{5}{*}{OMD}
                         &Feature Detection           &7.5220                              \cr
                         &Camera Pose Estimation      &32.0909                             \cr
                         &Dynamic Object Tracking (avg/object)    &7.0134                             \cr
                         &Object Motion Estimation   (avg/object) &19.5280                             \cr
                         &Map and Mask Updating       &30.3153                             \cr
                         &Local Batch Optimisation    &15.3414                \cr
  \bottomrule
 \end{tabular}
 % \vspace{-.2cm}
\end{table}

%%%%%%%%%%%%%%%%%%%%%%%%%%%%%%%%%%%%%%%%%%%%%%%%%%%%%%%%%%%%%%%%%%%%%%%%%%%%%%%%%%%%%%%%%%%%%%%%%%%%%%%%%%%%%%%

\section{Conclusion}
\label{sec:conclusion}
In this paper, we have presented VDO-SLAM, a novel dynamic feature-based SLAM system that exploits image-based semantic information in the scene with no additional knowledge of the object pose or geometry, to achieve simultaneous localisation, mapping and tracking of dynamic objects.
The system consistently shows robust and accurate results on indoor and challenging outdoor datasets, and achieves state-of-the-art performance in object motion estimation. We believe the high performance accuracy achieved in object motion estimation is due to the fact that our system is a feature-based system. Feature points remain to be the easiest to detect, track and integrate within a SLAM system, and that require the front-end to have no additional knowledge about the object model, or explicitly provide any information about its pose.  

%Further extension can be considered to apply adaptive motion models for objects in different scenarios.Learning-based methods can provide important cues to determine if a vehicle is accelerating, moving with a constant motion or decelerating based on attention models to street signs, traffic lights, etc.
An important issue to be reduced is the computational complexity of SLAM with dynamic objects.
In long-term applications, different techniques can be applied to limit the growth of the graph (\cite{Strasdat11iccv, Ila10tro}). In fact, history summarisation/deletion of map points pertaining to dynamic objects observed far in the past seems to be a natural step towards a long-term SLAM system in highly dynamic environments. 
%An important question is the map representation in dynamic environments.
%While we have decided to represent the environment as $3$D points, and believe this is at the core of the results we have achieved in object motion estimation, a richer representation (e.g., cuboid, quadric, etc.) could provide a more useful map for interaction with the dynamic world such as manipulation tasks, obstacle avoidance, etc$\dots$ An interesting future work would be a system that is able to switch between different map representations depending on the task performed.
%

%%%%%%%%%%%%%%%%%%%%%%%%%%%%%%%%%%%%%%%%%%%%%%%%%%%%%%%%%%%%%%%%%%%%%%%%%%%%%%%%%%%%%%%%%%%%%%%%%%%%%%%%%%%%%%%

\section*{Acknowledgements}
\label{sec:acknowledge}
This research is supported by the Australian Research Council through the Australian Centre of Excellence for Robotic Vision (CE140100016), and the Sydney Institute for Robotics and Intelligent Systems. The authors would like to thank Mr. Ziang Cheng and Mr. Huangying Zhan for providing help in preparing the testing datasets.

\ifCLASSOPTIONcaptionsoff
  \newpage
\fi



%\bibliographystyle{IEEEtran}
\bibliographystyle{IEEEtran}
\bibliography{bibliography}


% \end{IEEEbiography}


% % insert where needed to balance the two columns on the last page with
% %\newpage



%\vfill


\end{document}